\documentclass[lettersize,journal]{IEEEtran}
\usepackage{amsmath,amsfonts}
\usepackage{algorithmic}
\usepackage{algorithm}
\usepackage{array}
\usepackage[caption=false,font=normalsize,labelfont=sf,textfont=sf]{subfig}
\usepackage{textcomp}
\usepackage{stfloats}
\usepackage{url}
\usepackage{verbatim}
\usepackage{graphicx}
\usepackage{cite}
\usepackage{times}
\usepackage{epsfig}
\usepackage{amsmath}
\usepackage{amssymb}
\usepackage{multirow}
\usepackage{booktabs}
\usepackage{comment}
\usepackage[percent]{overpic}
\usepackage{tabularray}
\usepackage{xcolor}
\usepackage{stackengine}
\usepackage{afterpage}
\usepackage{dsfont}

\usepackage{hyperref}
\usepackage{bbding}
\hypersetup{
    colorlinks=true,
    linkcolor=magenta,
    filecolor=magenta,      
    urlcolor=cyan,
    pdftitle={Overleaf Example},
    pdfpagemode=FullScreen,
    }

\title{You Only Train Once:\\A Unified Framework for Both Full-Reference and No-Reference Image Quality Assessment}
\author{Yi~Ke~Yun,~\IEEEmembership{Student,~IEEE} and Weisi~Lin,~\IEEEmembership{Fellow,~IEEE}
}

\begin{document}


\maketitle

\begin{abstract}
Although recent efforts in image quality assessment (IQA) have achieved promising performance, there still exists a considerable gap compared to the human visual system (HVS). One significant disparity lies in humans' seamless transition between full reference (FR) and no reference (NR) tasks, whereas existing models are constrained to either FR or NR tasks. This disparity implies the necessity of designing two distinct systems, thereby greatly diminishing the model's versatility. Therefore, our focus lies in unifying FR and NR IQA under a single framework.
Specifically, we first employ an encoder to extract multi-level features from input images. Then a Hierarchical Attention (HA) module is proposed as a universal adapter for both FR and NR inputs to model the spatial distortion at each encoder stage. Furthermore, considering that different distortions contaminate encoder stages and damage image semantic meaning differently, a Semantic Distortion Aware (SDA) module is proposed to examine feature correlations between shallow and deep layers of the encoder. By adopting HA and SDA, the proposed network can effectively perform both FR and NR IQA. 
When our proposed model is independently trained on NR or FR IQA tasks, it outperforms existing models and achieves state-of-the-art performance. Moreover, when trained jointly on NR and FR IQA tasks, it further enhances the performance of NR IQA while achieving on-par performance in the state-of-the-art FR IQA. You only train once to perform both IQA tasks. Code will be released at: \textcolor{magenta}{https://github.com/BarCodeReader/YOTO}.

\end{abstract}

\begin{IEEEkeywords}
Full Reference IQA, No Reference IQA, Transformer, Unified Structure
\end{IEEEkeywords}

\section{Introduction}
The importance of accurately assessing image quality has become increasingly crucial in many fields \cite{survey1, survey2, traffic_sign_detection, traffic_control, tres}. Though the human vision system (HVS) is capable of identifying high-quality images effortlessly, it is labor-intensive, and in most cases infeasible, to assess image quality via human workers. Therefore, image quality assessment (IQA) aims to develop objective metrics that can predict image quality as perceived by humans \cite{AMIQ}.

\begin{figure}[!t]
\centering
\includegraphics[width=\linewidth]{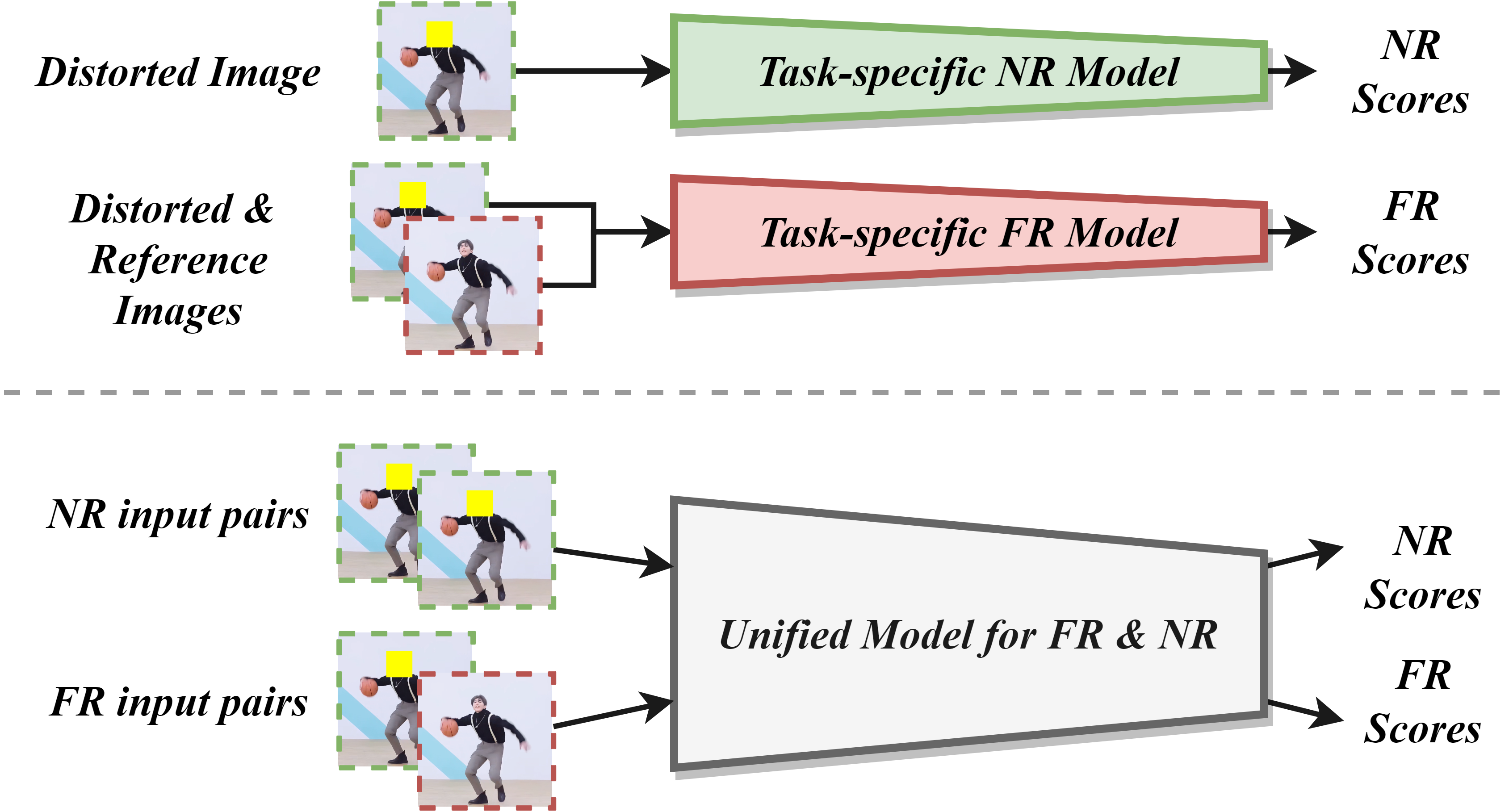}
\caption{Illustration of the main purpose for the proposed network (the lower part in the figure), where different image pairs can be fed into the network to yield FR/NR IQA scores using the same architecture. The traditional framework is given as the upper part of the figure for easy comparison. Our method offers great simplicity in both training and applications, minimizing performance inconsistencies on switching FR/NR tasks, and it achieves state-of-the-art performance on both FR and NR IQA benchmarks.}
\label{figure1}
\end{figure}

Depending on the presence of high-quality reference images, IQA tasks can be grouped into two categories: No-Reference (NR) and Full-Reference (FR) \footnote{In some papers IQA with partial reference, Reduced-Reference (RR), is introduced as the third category.}. Early studies apply hand-crafted metrics to perform the IQA. FR-IQA metrics like PSNR, SSIM \cite{SSIM}, and NSER \cite{NSER} assess image quality via signal fidelity and structural similarity between distorted and reference images. NR-IQA metrics such as DIIVINE \cite{diivine} and BRISQUE \cite{brisque} attempt to model natural scene statistics for quality measurement. These methods are well-defined for specific distortions and are usually poorly generalized. Unlike other methods that compare distorted images with pristine images, BPRI \cite{BPRI} and BMPRI \cite{BMPRI} aim to first obtain characteristics from the image that suffers the most distortion, namely pseudo-reference image (PRI), then estimate quality scores by the blockiness, sharpness, and noisiness between distorted images and PRIs.

In the deep learning era, the core idea for most FR-IQA methods is to estimate quality score by identifying low-quality regions through differencing the reference and distorted images or their extracted features \cite{DEEPQA, DUALCNN, PIEAPP}. For NR-IQA, common practices are adopting ranking \cite{rankiqa, ranking_iqa, dacnn}, feature comparison \cite{contrast, OHEM}, generating pristine images using Generative Adversarial Networks (GAN) \cite{HALUIQA, BIQA_AI, GANIQA, GAN_TIP}, and applying Transformers or multi-scale CNNs for better spatial modeling of distortions \cite{stairiqa, tres, maniqa, multiscale, RL_ATTN, view_tf, cascade_iqa}. The rationale behind these NR methods is to implicitly identify low-quality regions for IQA score estimation. 

\begin{figure}[!t]
\centering
\begin{overpic}[width=0.24\linewidth]
{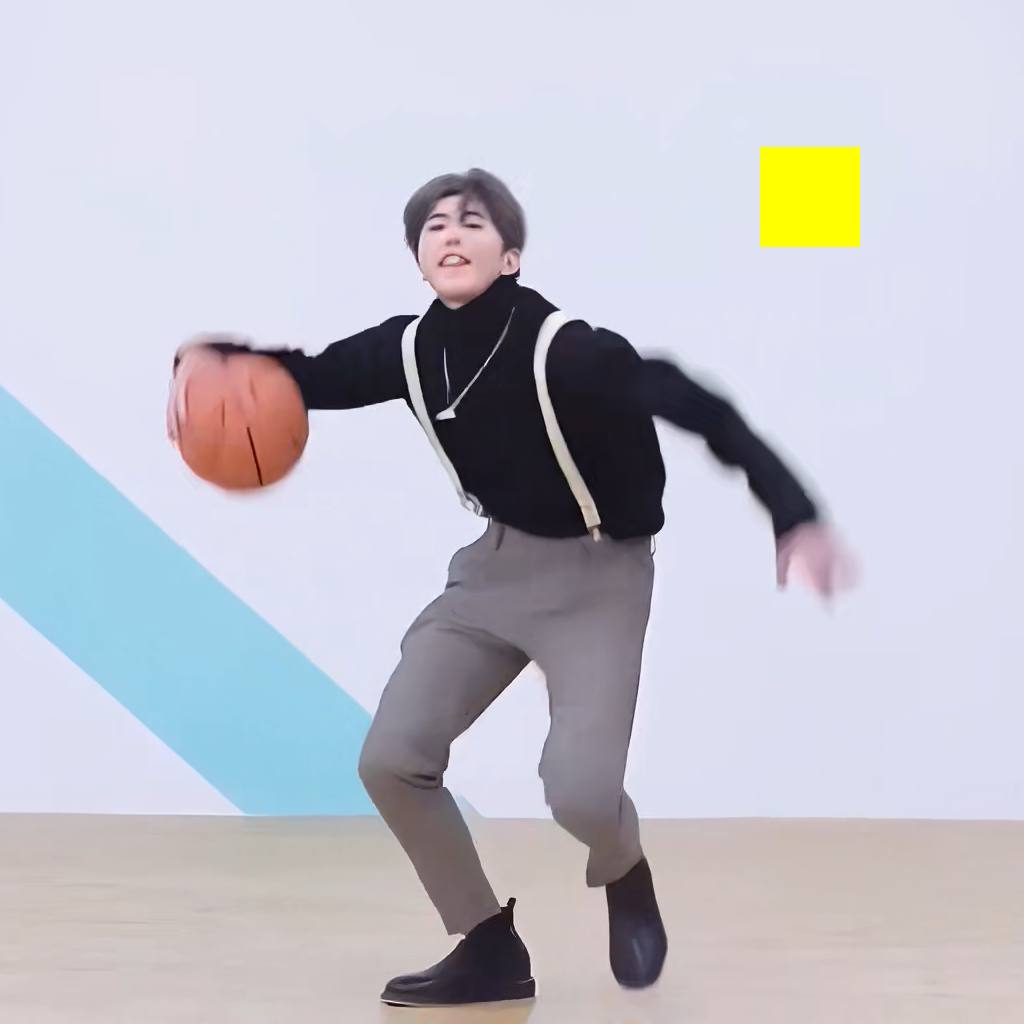}
\put (5,90) {\textcolor{black}{(a)}}
\end{overpic}
\begin{overpic}[width=0.24\linewidth]
{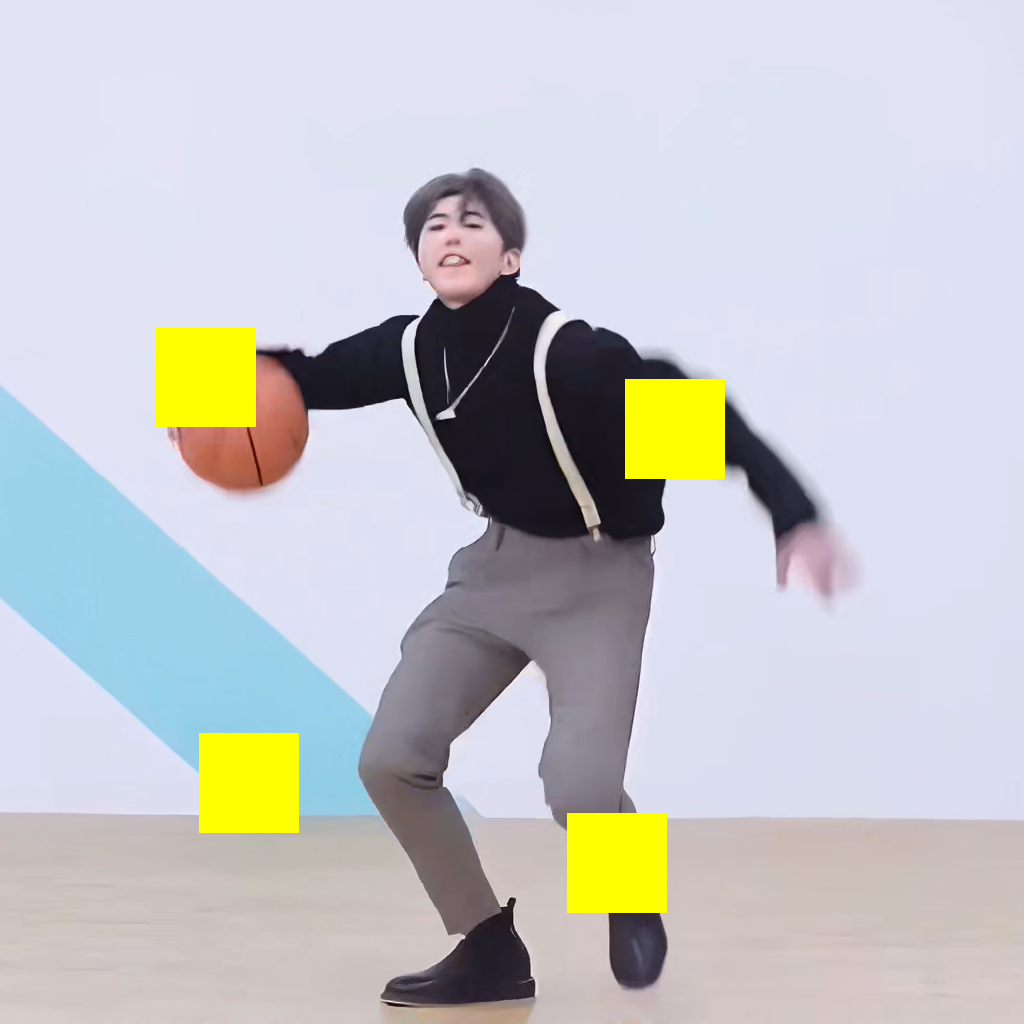}
\put (5,90) {\textcolor{black}{(b)}}
\end{overpic}
\begin{overpic}[width=0.24\linewidth]
{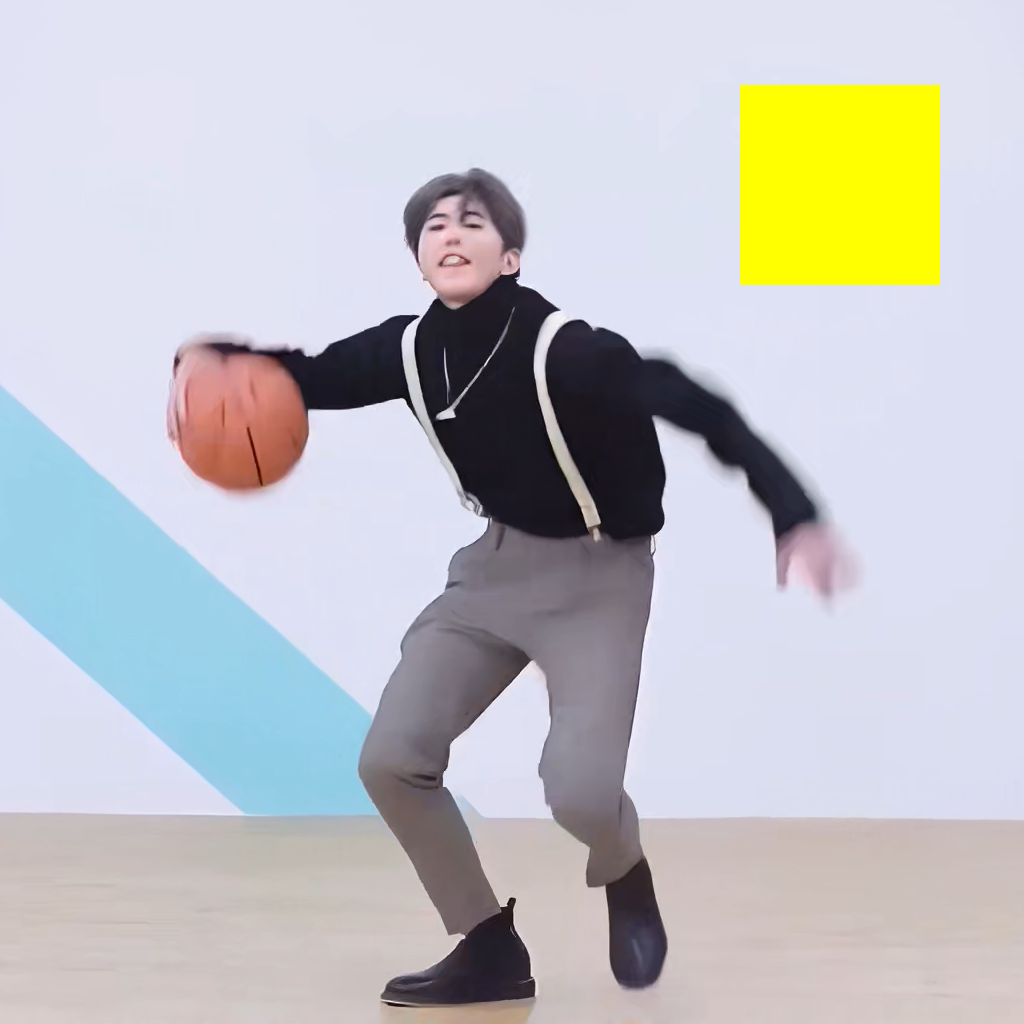}
\put (5,90) {\textcolor{black}{(c)}}
\end{overpic}
\begin{overpic}[width=0.24\linewidth]
{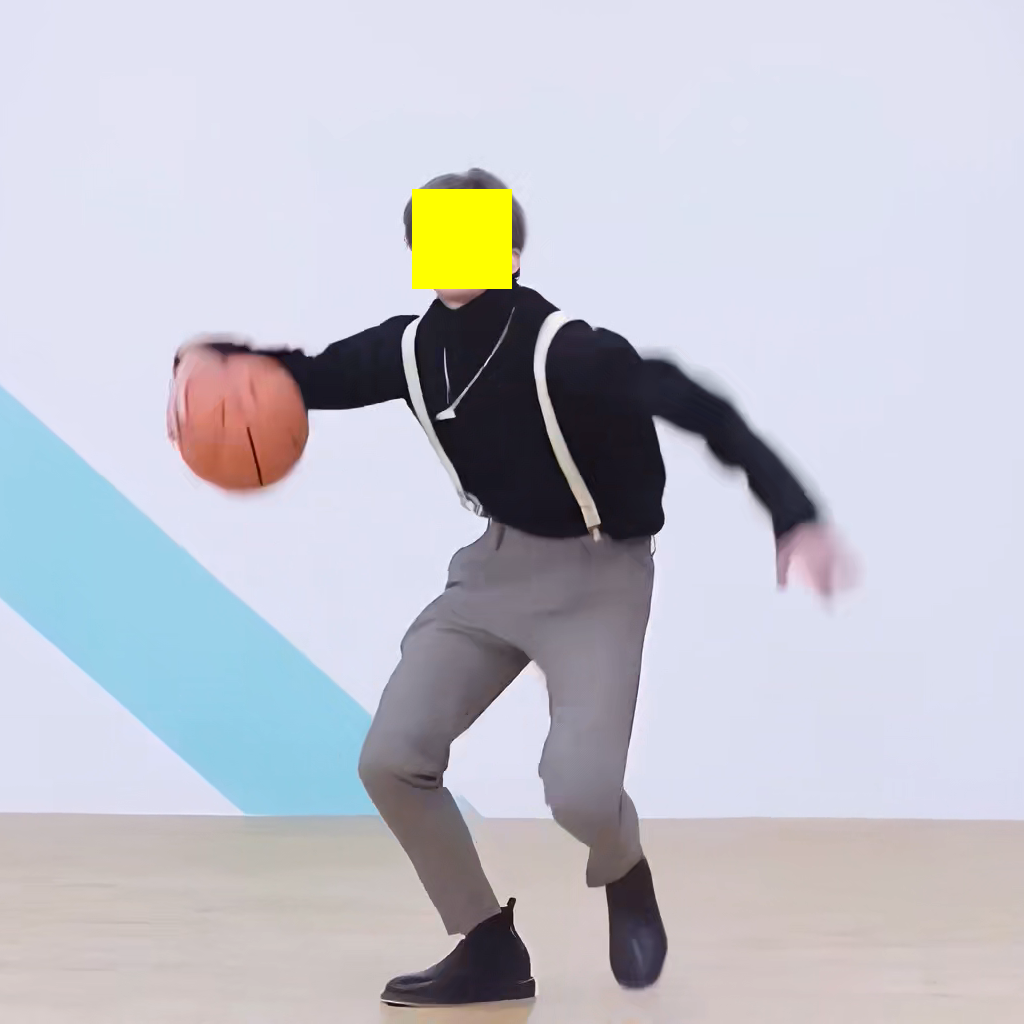}
\put (5,90) {\textcolor{black}{(d)}}
\end{overpic}
\caption{Illustration of how humans assess image quality when distortion is present: which one has the highest quality score? (a) a $100\times100$ yellow block is presented in the background. (b) four blocks in both background and foreground (the basketball player). (c) a $200\times200$ block in the background. (d) a $100\times100$ block on the face. Quality score based on the amount of distortion: $(b)<(a)$ and $(c)<(a)$, while that based on the semantic impact to content (the player): $(b)<(c)$ and thus $(b)<(c)<(a)$. How about (d)? It has less amount of distortions than (b) and (c). However, it should have the lowest quality score because the most important message, the face, is damaged. Thus, aside from the amount, distortion is significant if it has critical damage to an image's semantic meaning.}
\label{kunkun}
\end{figure}

With or without the reference image, human beings can perform both FR and NR IQA effortlessly. This indicates that the FR and NR IQA share commonalities and there is a general model in our brains capable of performing both tasks. Recent research has found Saliency Map \cite{JNDSAL, sgdnet, WADIQAM, sal_iqa2} to be beneficial for both FR and NR IQA tasks. This serves as additional evidence demonstrating the commonalities between FR and NR IQA. Unfortunately, methods in the field are typically task-specific designed, as illustrated in Fig.\ref{figure1}. They are indeed effective; however, once trained, they cannot be readily transferred to other IQA tasks (like from NR to FR or vice versa). Consequently, they lack a certain degree of generality and still exhibit a gap compared to humans, thus hindering our exploration of the underlying essence of IQA. A model capable of performing both IQA tasks can narrow this gap and thus it is meaningful and necessary. Besides, not only are human beings capable of perceiving the amount of distortion present in an image, but we are also adept at identifying the extent to which such distortion affects the semantic meaning, as illustrated in Fig.\ref{kunkun}. Though the field has recognized the significance of semantic information such as saliency, we lack an effective method for modeling the impact of distortion on semantic meaning.

To this end, we aim to narrow the gap between FR and NR IQA by developing a unified model and to improve existing IQA performance from the perspective of semantic modeling of distortion.
Specifically, we first apply an encoder (ResNet50 \cite{RESNET} or Swin Transformer \cite{SWIN}) for feature extraction.  Input images are encoded into multi-scale features. We then utilize the attention mechanism as the universal adaptor for FR and NR tasks. Depending on the input pairs, i.e. $\{distorted\ img.,\ distorted\ img.\}$ or $\{distorted\ img.,\ pristine\ img.\}$, the attention block can dynamically switch between self-attention and cross-attention thus unifying FR and NR tasks. However, global attention is usually sparse due to the large spatial dimension of shallow features. To further model the spatial distortion, by applying a scale factor $r$, we partitioned the attention matrix into patches and computed the local attention. We stack several attention blocks together with different scale factors and name it the Hierarchical Attention (HA) module. To model the semantic impact caused by distortion, we developed a Semantic Distortion Aware (SDA) module. Distortions at one encoder stage, based on their types and strengths, might remain and affect the next few encoder stages differently. Though self-attention is effective in identifying spatially spread distortions, it has a weak representation of how distortions are correlated across all encoder stages. To this end, we propose to compute the cross-attention between features from different encoder stages. To ensure efficiency, the calculation is performed only between high-level feature pixels and their associated low-level sub-regions (like a "cone"). Lastly, features from HA and SDA are concatenated and aggregated for quality score generation.

\begin{figure}[!t]
\centering
\includegraphics[width=\linewidth]{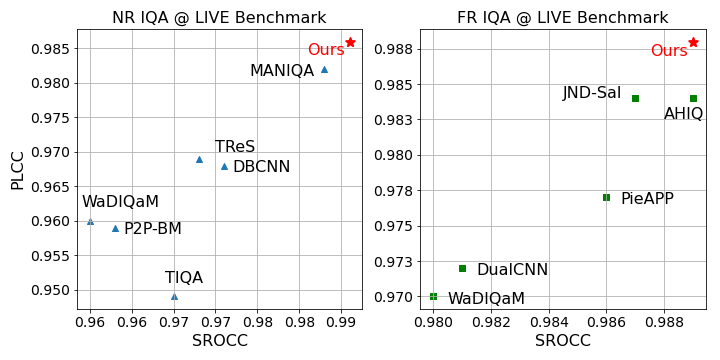} 
\caption{Performance comparison against other FR and NR IQA models on LIVE \cite{live_dataset} dataset. Our method achieves state-of-the-art performance on both FR and NR IQA benchmarks using the same network architecture.}
\label{performance}
\end{figure}

Extensive experiments on four FR-IQA benchmarks and seven NR-IQA benchmarks demonstrate the effectiveness of our proposed network. The proposed network outperforms other state-of-the-art methods on both FR and NR tasks using a unified architecture, as shown in Fig.\ref{performance}. To sum up, our contributions are threefold as follows:
\begin{itemize}
    \item We propose a unified framework, YOTO, for both FR and NR IQA tasks and achieve state-of-the-art performance. The unified structure offers great simplicity and to our best knowledge, we are the first to aim to complete these two tasks using the same network architecture 
    for IQA.
    \item We introduce the Hierarchical Attention (HA) module as the universal adapter for both NR and FR inputs and for spatial distortion modeling. We further devise a Semantic Distortion Aware (SDA) module for semantic impact modeling.
    \item The unified network also enables joint training on both FR and NR IQA. \textbf{You only train once}, and the obtained model can achieve on-par performance on the FR task and better performance on the NR task trained separately. 
\end{itemize}

\section{Related Work}

\begin{figure*}[!t]
\centering
\includegraphics[width=\linewidth]{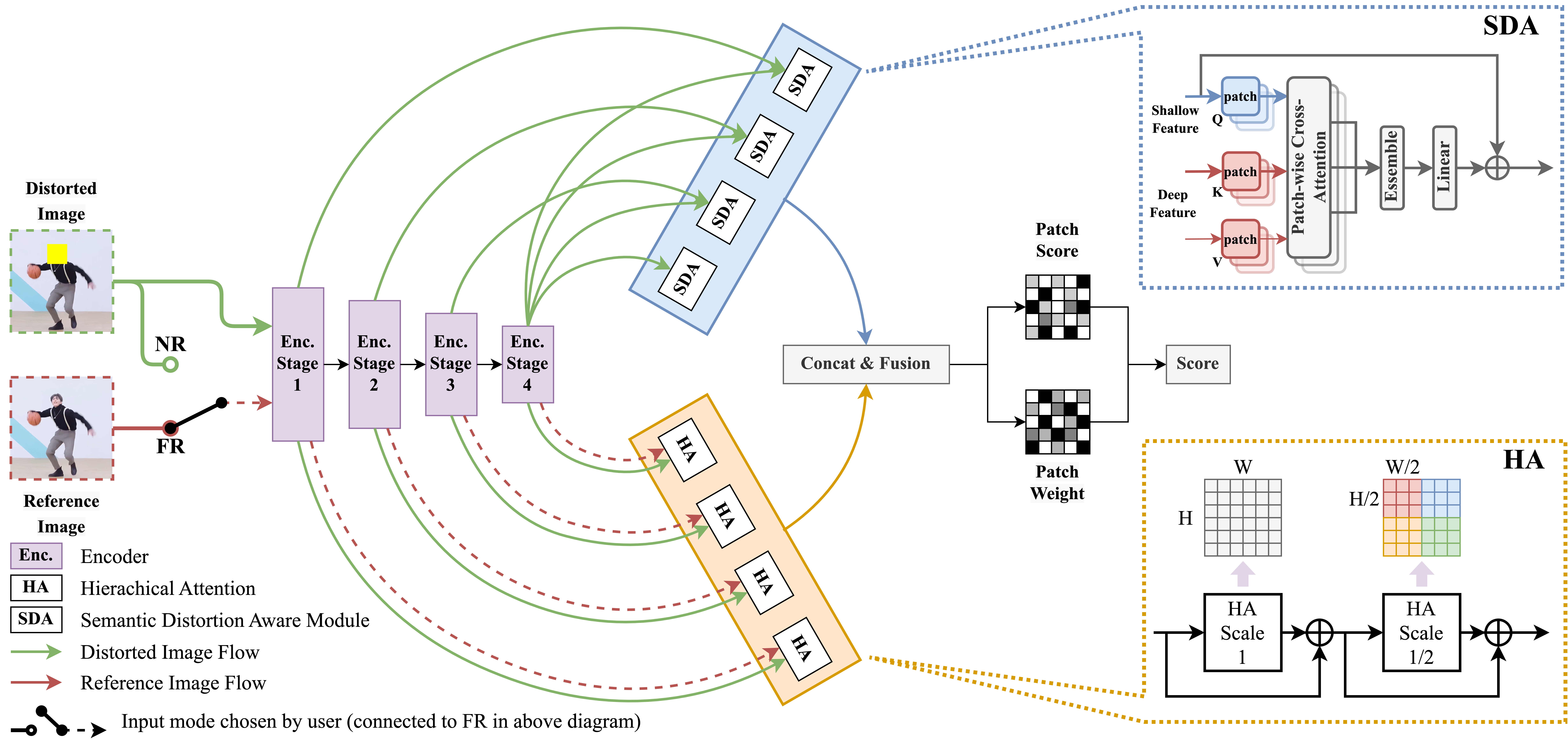} 
\caption{Network architecture of the proposed YOTO. Input types can be chosen by the user depending on the presence of reference images for the network to perform FR or NR IQA tasks (i.e. the dotted red arrow will become green if the user chooses to perform NR IQA). The network receives a pair of images in the form of $[distorted, distorted]$ or $[distorted, reference]$ for NR and FR IQA respectively. ResNet50 or Swin Transformer is adopted as the encoder backbone (purple). A Hierarchical Attention (HA) module with global and regional attention is developed to highlight potential distortion-contaminated areas in encoder features. If only distorted images are provided, self-attention is applied in the HA module. If reference images are given, the HA module will compute the cross-attention between distorted and reference features. To model the semantic impact caused by distortion, a Semantic Distortion Aware (SDA) module is designed and densely applied to explore the similarity between shallow and deep features using only distorted image features. The obtained features from HA and SDA are concatenated and fused for IQA score estimation via a commonly applied patch-wise attention. Best viewed in color.}
\label{YOTO}
\end{figure*}

\textbf{FR-IQA} Since reference images are given alongside distorted images, the core idea is to estimate the quality scores based on the differences. The most straightforward way is to calculate the Mean-squared Error (MSE) between distorted images and reference images. However, MSE cannot effectively reflect visual quality perceived by humans \cite{MAE_WRONG} and various image quality metrics have been developed striving for better alignment with HVS \cite{IQA_SURVEY}. With the emergence of deep learning, DeepQA \cite{DEEPQA} applies CNN on distorted images and their corresponding difference maps to estimate quality scores. Similarly, DualCNN \cite{DUALCNN} and PieAPP \cite{PIEAPP} perform IQA based on the feature differences between pristine images and distorted images. These methods focus on modeling spatial differences but ignore the semantic impacts caused by distortion. To mitigate this issue, saliency was introduced to reflect visually important regions \cite{JNDSAL,SAL_FUSION,SAL_IQA,SAL_PRIOR}. However, saliency is simply served as an attention map and usually requires an extra saliency network for saliency generation. Thus better modeling of semantic impact caused by distortion still awaits more investigation.

\textbf{NR-IQA} One of the ideas for NR IQA is to generate the reference image using GAN-based methods \cite{HALUIQA, BIQA_AI} to transform NR into FR tasks. However, the performance of this approach is greatly bottlenecked by the GAN itself. Without explicitly generating reference images, similar to FR IQA, saliency \cite{sgdnet} is usually considered as auxiliary information for better quality score estimation. Alternatively, ranking \cite{rankiqa, MISTRIQ} is commonly applied to implicitly learn the distortion level in images using Siamese networks. Besides, Vision Transformers \cite{tres, maniqa, tiqa, IQT, MISTRIQ} are adopted to capture better spatial relationships aiming to identify hidden distortion patterns. Although the aforementioned methods share great similarities with FR IQA, most networks are typically designed only for NR tasks. An ideal model should be able to perform both FR and NR IQA tasks, just like human beings. To our best knowledge, the only method \cite{WADIQAM} reporting performances on both NR and FR benchmarks still requires modifications of its network architecture, and its performance is inferior to the latest single-task models. Hence, the exploration of a unified model is non-trivial, as it can narrow the gap between current models and human capabilities, while also inspiring and facilitating a better exploration of underlying principles between FR and NR IQA.

\textbf{Cross-attention} Cross-attention was originally introduced in Natural Language Processing (NLP) to translate a source language to another (eg., English to French) \cite{NLP_attn}. Due to its ability to examine correlations between different modalities, it has been widely adopted in various NLP and computer vision applications, such as text and image matching \cite{CA_TEXTIMG_MATCH, CA_TEXTIMG_MATCH2}. It is also widely applied in Transformers for multi-scale image feature correlation modeling \cite{crossVIT, CAT}. For IQA tasks, various self-attention-related models have been proposed thanks to the long-range dependency modeling capability of transformers \cite{AHIQ, maniqa, IQT, MISTRIQ}. The self-attention is applied at each encoder stage and the resulting weighted features are effective in identifying distortion in spatial dimensions. However, part of the distortion at one encoder stage might remain and continue to affect the next few encoder stages. This part of features should gain higher attention consistently as they have greater impacts on the semantics of the image. Thus solely applying self-attention for each encoder stage cannot effectively reflect the semantic impact caused by distortion. Cross-attention, on the other side, is not well investigated yet and seems a promising solution for semantic impact modeling. Since different types of distortion will contaminate encoder layers differently, we can apply cross-attention to calculate correlations between shallow features and deep features to estimate the strength of distortion affecting the semantic meaning of an image.

\section{Proposed Method}
\subsection{Overall Architecture}
The overall architecture is shown in Fig.\ref{YOTO}. We adopt ResNet \cite{RESNET} or Swin Transformer \cite{SWIN} as our encoder backbone for image feature encoding. The encoder is shared if reference images are present. Channel-wise attention is applied to encoder features. After that, a Hierarchical Attention (HA) module is developed to be an adaptor for NR and FR inputs and to identify potentially damaged regions in encoder features. The self-attention employed in the HA module will become cross-attention if pristine image features are present. Therefore the network is capable of receiving both FR and NR inputs. Furthermore, a Semantic Distortion Aware (SDA) module is proposed to estimate the semantic impact caused by distortion presented in the image. Finally, an aggregation block with patch-wise attention is devised for quality scores.

\subsection{Hierarchical Attention (HA)}
In order to simultaneously address FR and NR IQA tasks, a network requires an adaptive input processing module. The module should be capable of dynamically adjusting the calculation based on the presence or absence of reference images. In this regard, the attention mechanism from the original Transformer \cite{NLP_attn} is considered as a potential solution. When only distorted images are available, self-attention can be utilized to enable the network to identify the possible locations of distortion and adjust the weight of features accordingly. On the other hand, if pristine images are provided in addition to distorted images, cross-attention can be employed to achieve the same objective without increasing computational complexity. Denote the distorted features and reference features as subscripts $dis$ and $ref$, the attention devised can be described as:
\begin{equation}
    Attention= \begin{cases}
    Softmax(\frac{Q_{dis}K_{dis}^T}{\sqrt{d_{head}}})V_{dis}, & \text{if $NR$}.\\
    Softmax(\frac{Q_{ref}K_{dis}^T}{\sqrt{d_{head}}})V_{dis}, & \text{if $FR$}.
  \end{cases}
\end{equation}
where $Q,K,V$ are embeddings calculated by different MLP layers. 

\begin{figure}[!t]
\centering
\includegraphics[width=\linewidth]{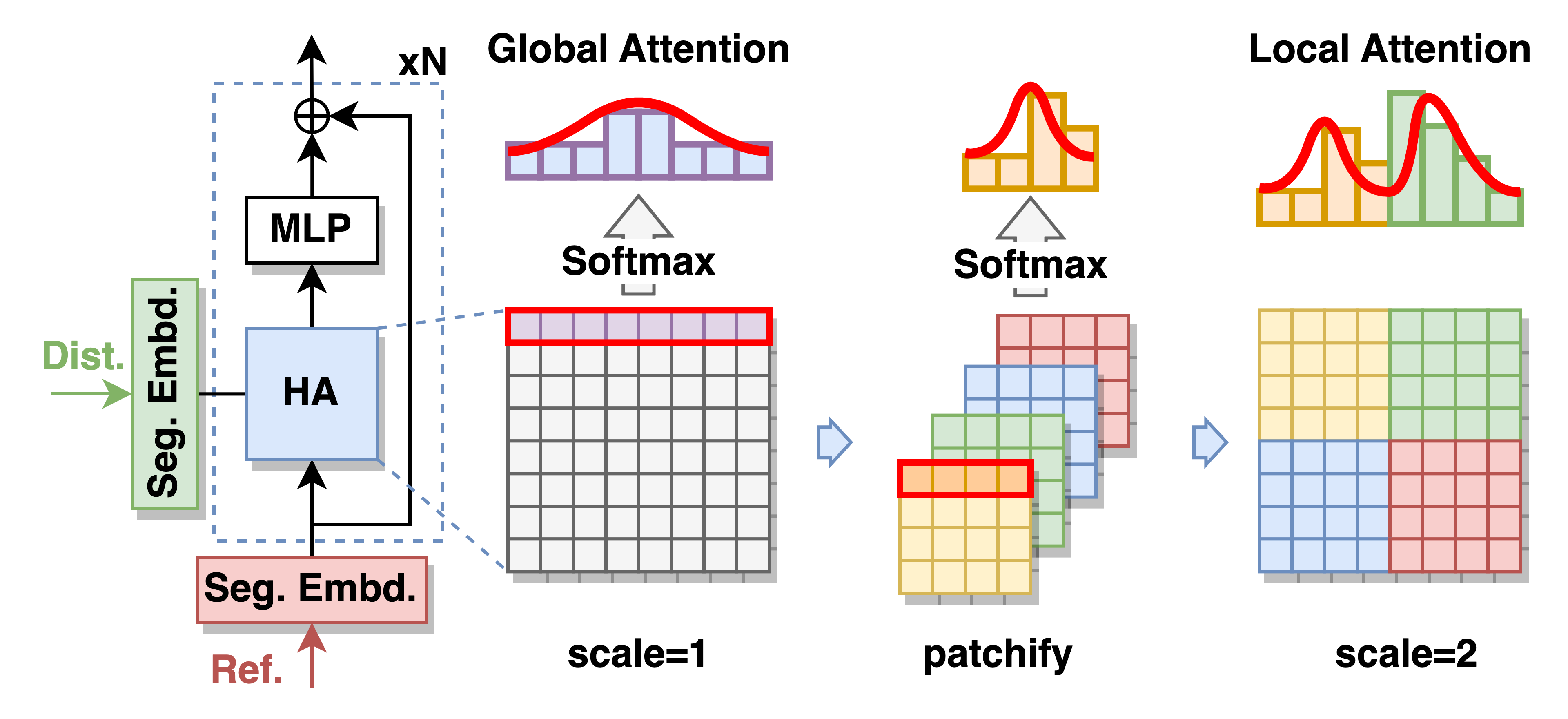} 
\caption{The developed Hierarchical Attention (HA) module that bridges FR and NR tasks together. Segmentation embeddings were added to input features as an indicator of NR and FR tasks. Depending on FR or NR tasks, the input pair shown above could be $\{distortion, reference\}$ or $\{distortion, distortion\}$ respectively. Besides, on top of global attention, we partitioned the attention matrix into patches to incorporate local attention as well. In practice, the HA module will stack multiple attention layers with different scale factors to fully exploit global and local attention.}
\label{HA}
\end{figure}

However, the network has no idea which and when the pristine image is present in the input pairs. To this end, we introduce a segmentation embedding layer $embd(\cdot)$ indicating the input information of FR and NR IQA, as shown in Fig.\ref{HA}.
After embedding, with the developed HA, the network can now perform both NR and FR tasks automatically based on different input image pairs:
\begin{equation}
    score = \begin{cases}
    \mathcal{F}(embd(\textbf{0})+x_{dis}, embd(\textbf{0})+x_{dis}), & \text{if $NR$}.\\
    \mathcal{F}(embd(\textbf{0})+x_{dis}, embd(\textbf{1})+x_{ref}), & \text{if $FR$}.
  \end{cases}
\end{equation}
where $x_{dis}$ and $x_{ref}$ denotes distorted and reference images, $\mathcal{F}$ stands for the network and $\textbf{\{0, 1\}}$ are embedding vectors.

The adoption of attention resolves the issue of different inputs for FR and NR IQA. When computing attention on the encoder features, the attention matrix formed by patch embeddings can become quite large due to the typically large spatial dimension of shallow features. For example, a $512\times512$ single-channel image split into non-overlapping patches of $16\times16$ will give a sequence of $32\times32=1024$ in length and result in an attention matrix of $1024\times1024$. As a result, after applying softmax to each row, the attention scores between patches might be very similar, leading to an indistinct weighting effect on the features. NLP and ViT models commonly resolve this problem by stacking multiple attention layers. However, this approach may not be efficient in terms of computation cost and convergence speed. To this end, we partition the feature maps into blocks and compute attention for each block separately, and then concatenate the resulting attention matrices into a large matrix. \textbf{Please note our approach has a distinct difference from Multi-head Self-attention (MSA) \cite{NLP_attn} as we split features from shape $[Batch, Length, Channels]$ to $[Batch, scale^2, Length/scale^2, Channels]$ while in MSA the channel dimension is split}. As we can choose different scale factors to divide the original matrix into various sizes of blocks, we name the designed module Hierarchical Attention (HA). With this approach, both global and local attention are exploited. Mathematically, given an input sequence $x\in R^{B\times HW\times C}$, a scale factor $r$ will split the attention matrix $M \in R^{B\times HW\times HW}$ into $r^2$ blocks $M_{ij}$ where $i,j\in[0,r)$. Defining the operation $\displaystyle P_{i,j=0}^{r-1} M_{ij}$ to paste the attention matrix $M_{ij}$ back to the original matrix $M$ following its coordinate $[i,j]$, a single layer of HA can be calculated as:
\begin{equation}
Q_{r,ij},K_{r,ij} \in R^{B\times (r)^2 \times HW/(r)^2\times C}
\end{equation}
\begin{equation}
M_{ij} = Softmax(\frac{Q_{r,ij}K_{r,ij}^T}{\sqrt{d_{head}}})
\end{equation}
\begin{equation}
M = P_{i,j=0}^{r-1} M_{ij}
\end{equation}
\begin{equation}
HA(x_t,r) = x_t + MLP(MV), \ \ \ V\in R^{B\times HW\times C}
\end{equation}
The HA module can be described recursively:
\begin{equation}
HA(x_{t+1}, r) = HA(HA(x_t, r), r), \ \ \ t = 0, 1, ... 
\end{equation}

The overall design of the HA module is illustrated in Fig.\ref{HA} where we stack multiple HA layers with different scale factors and apply residual connections. The effectiveness of the HA module will be discussed more in ablations.

\subsection{Semantic Distortion Aware (SDA) Module}

The HA module provides effective attention allowing the model to focus on certain regions of the image. However, as mentioned in the previous sections, relying solely on the amount of distortion in the image may not adequately reflect image quality as we ignored the impact distortion applied on deeper encoder layers. Thus, we need to assess the damage to the semantic information caused by distortion for better evaluation. Considering that the encoder itself is a natural feature extractor or feature summarizer, i.e., the details and textures in the image are gradually filtered out during the encoding process, leaving behind high-level semantic information such as location. Based on this characteristic, we propose to evaluate the amount of semantic impact caused by different types of distortion by examining their residual effects in each encoder layer. Different from common approaches \cite{hyperiqa, tres, metaiqa} where features are aggregated or scores are averaged, we aim to estimate the semantic damage caused by the distortions from the perspective of feature consistency. The distortion that affects the next few stages of the encoder should receive more attention. For instance, we believe the impact of salt-and-pepper distortion and block occlusion will reach different layers when they pass through the encoder. Salt-and-pepper distortion is likely to be filtered out effectively in the first few layers of the encoder, while block occlusion may penetrate till the last layer of the encoder, affecting the semantics of the entire image.

\begin{figure*}[!t]
\centering
\includegraphics[width=\linewidth]{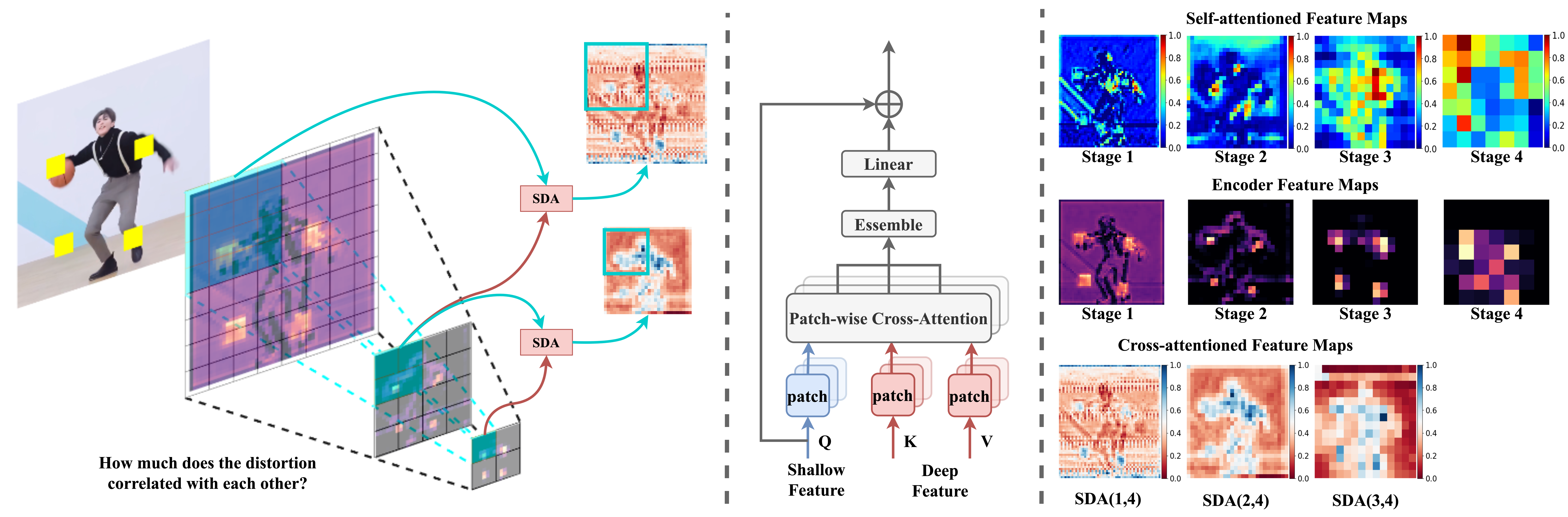} 
\caption{Illustration of SDA (left and middle) and comparison against self-attention (right). Feature pyramid is firstly split into $n\times n$ cones. The SDA is then densely applied on encoder stages to explore the correlation between shallow and deep features so that we can estimate the semantic impact caused by distortion. SDA(1,4) denotes the weighted feature map obtained by computing cross-attention between stages 1 and 4. As illustrated on the right side, the cross-attention feature maps (last row) can reflect how distortions in shallow stages are related to the deep features, further indicating the impact the distortion caused. Self-attention features (first row), on the other side, indicate the important region in space but are inconsistent across different encoder stages and thus have a weak representation of semantic impact.}
\label{SDA}
\end{figure*}

We propose using cross-attention between different encoder layers to capture the impact of distortion on semantic meaning. By comparing the feature similarities between shallow and deep layers, we estimate the distortion residues in different layers and evaluate image quality. Therefore, we design the Semantic Distortion Aware (SDA) module, as shown in Fig.\ref{SDA}. It is a cross-attention module in a top-down manner and is densely applied between different encoder features. At first glance, the proposed SDA might look superficially simple. \textbf{In fact, instead of blindly applying the cross-attention, we partition the feature pyramid into a "cone" and trace a local patch from shallow to deep features to estimate the semantic correlations.} This design prevents meaningless interactions between a local distortion and its high-level neighbor regions during cross-attention. It bridges low-level and high-level features, focusing on how local distortions within low-level features affect high-level semantics. Meanwhile, it is more computationally efficient ($1/n^{2}$ if has $n\times n$ cones) compared with full-scale attention. The SDA module uses shallow features as queries $Q_s$ and deep features as keys $K_d$ and values $V_d$ to calculate the weighted map of shallow-layer features, reflecting which region on the image has a direct correlation and persistent effect on its deep semantic meanings. Given shallow and deep features $x_s, x_d$ with batch $B$, channel $C$ and spatial dimension $m\times m$ and $n\times n$, then we have:
\begin{equation}
    \begin{gathered}
        x_s \in R^{B\times (m\times m)\times C},\ \ \ x_d \in R^{B\times (n\times n)\times C}, m>n\\
        Attention(Q_s, K_d, V_d) = Softmax(\frac{Q_{s}K_{d}^T}{\sqrt{d_{head}}})V_{d} \\
        SDA(x_d, x_s) = x_s+ MLP(Attention(Q_s, K_d, V_d))
    \end{gathered}
\end{equation}

For each encoder stage $1-4$, we apply HA to obtain features $H_1, H_2, H_3$, and $H_4$ accordingly. Similarly, we densely apply SDA between encoder stage $i$ and its shallower stages j $(j<i)$ to obtain correlation features, $C_{i,j}$. This will yield in $C_{4,3}, C_{4,2},..., C_{2,1}$ in total six feature groups. The obtained features from the same encoder stage $i$ are averaged and concatenated along the channel dimension followed by a commonly applied patch-wise attention block \cite{maniqa, AHIQ, WADIQAM} to predict quality scores, as shown in Fig.\ref{YOTO}.

\subsection{Training Schemes}
Due to the proposed model architecture's ability to dynamically transform computation methods by accepting FR and NR IQA inputs, we have explored the possibility of conducting joint training for FR and NR alongside traditional task-specific training. FR and NR inputs are provided to the model with a probability of 50\% in a format of $\{distortion, reference\}$ or $\{distortion, distortion\}$. \textbf{We found that, compared to separate training, joint training not only achieves similar performance in FR but also significantly enhances NR performance.} Further details will be presented in the experiments and ablation analyses.

\section{Experiments}
\subsection{Datasets and Metrics}
\label{dataset}
We evaluate our algorithm on conventional seven NR benchmark datasets and four FR benchmark datasets. In addition, we also report the results on the PIPAL\cite{PIPAL} dataset consisting of GAN-generated images. For NR benchmarks, four synthetic distorted datasets, TID2013 \cite{tid2013_dataset}, LIVE \cite{live_dataset}, CSIQ \cite{csiq_dataset}, and Kadid10k \cite{kadid10k_dataset} are adopted. These synthetic distorted datasets are also employed for FR evaluations. Besides, we use three authentic datasets, namely LIVE Challenge \cite{CLIVE}, KonIQ10k \cite{KONIQ}, and LIVEFB \cite{LIVEFB}, to further test our model's performance on NR tasks.

\textbf{TID2013}: TID2013 dataset contains 3000 distorted images derived from 25 pristine images with 24 different distortion types at 5 degradation levels. Mean Opinion Scores are ranging from 0 to 9.

\textbf{LIVE}: Live degrades 29 pristine images by adding additive white Gaussian noise, blurring Gaussian, compressing JPEG, compressing JPEG2000, and fast fading, resulting in 770 distorted images in total.

\textbf{CSIQ}: The dataset contains 886 distorted images originating from 30 pristine images. The CSIQ database includes six types of distortions, each with four or five levels of distortion respectively.

\textbf{Kadid10K}: This dataset contains 10,125 images that have been degraded based on 81 pristine images. Five levels of distortion are applied to each pristine image.

\textbf{KONIQ-10K}:
The KONIQ-10K dataset consists of 10,073 images covering a wide variety of images from various domains, including nature, people, animals, and objects. Each image in KONIQ-10K is rated on a scale from 1 to 10.

\textbf{LIVE-Challenge}:
The LIVE-Challenge dataset consists of 1,160 distorted images. The distortions include compression artifacts, noise, blurring, and other types of impairments.

\textbf{LIVE-FB}:
The LIVE-FB dataset is the largest in-the-wild NR-IQA dataset by far and it contains around 40 thousand distorted images and 1.2 million image patches labeled by crowd-sourcing.

\textbf{PIPAL}:
The PIPAL dataset introduces GAN-generated images as a new distortion type. Unlike conventional distortion types, the outputs from GAN-based models follow the natural image distribution quite well but with the wrong details. The PIPAL dataset consists of 250 pristine images with 40 distortion types and in total over 1 million labelings.

For metrics, we adopt Pearson’s Linear Correlation Coefficient (PLCC) and Spearman’s Rank Order Correlation Coefficient (SROCC) as our evaluation metrics. PLCC is defined as:
\begin{equation}
PLCC = \frac{{}\sum_{i=1}^{n} (x_i - \overline{x})(y_i - \overline{y})}{\sqrt{\sum_{i=1}^{n} (x_i - \overline{x})^2(y_i - \overline{y})^2}}
\end{equation}
where $x_i, y_i$ indicate the predicted and the ground truth scores, and their mean values are denoted as $\overline{x}, \overline{y}$ respectively. Meanwhile, SROCC is defined as: \begin{equation}
SROCC = \rho = 1- {\frac {6 \sum d_i^2}{n(n^2 - 1)}}
\end{equation}
where $n$ is the total number of samples, and $d_i$ is the rank difference of the $i^{th}$ test image between ground-truth and predictions.

\subsection{Implementation Details}
For the PIPAL dataset, we follow their official split and training protocols mentioned in the paper. For the other datasets mentioned in Sec.\ref{dataset}, we randomly sample 80\% of the data as the training dataset and the rest as the testing dataset. We choose ResNet50 and Swin Transformer as our encoder backbone to compare with other CNN and transformer models respectively. 
During training, images are augmented by random cropping with a size of $224\times 224$ and random horizontal flipping by 50\% chance. The learning rate is set to $1e^{-4}$ using the CosineAnneling scheduler with \textit{Tmax} and \textit{eta\_min} set to 50 and 0, respectively. We use a batch size of 8 and Adam optimizer \cite{adam} with default parameters. The network is trained using L2 loss for 100 epochs. During testing, each image is randomly cropped into $224\times 224$ for 20 times, and their averaged prediction scores are calculated. We train the network 5 times using different seeds and report the mean value. \textit{Code will be released upon acceptance.}

\subsection{Comparisons with the State-of-the-art}
\textbf{Single Task Training}
We compare our proposed YOTO against 13 NR IQA algorithms in the field, namely DIIVINE \cite{diivine}, BRISQUE \cite{brisque}, ILNIQE \cite{ilniqe}, BIECON \cite{biecon}, MEON \cite{meon}, WaDIQaM \cite{wdm}, DBCNN   \cite{dbcnn}, MetaIQA \cite{metaiqa}, P2P-BM \cite{p2pbm}, HyperIQA \cite{hyperiqa}, TReS \cite{tres}, TIQA \cite{tiqa}, and MANIQA  \cite{maniqa}. ResNet50 and Swin Transformer are used as our backbones respectively in order to compare the performance of other ResNet50-based and Transformer-based models. As listed in Tab.\ref{NR_compare}, our proposed network outperforms other algorithms on both ResNet-base and Transformer-based architectures.

Besides, we compare YOTO against 6 FR IQA methods, which are DOG \cite{DOG}, DeepQA \cite{DEEPQA}, DualCNN \cite{DUALCNN}, WaDIQaM\cite{WADIQAM}, PieAPP \cite{PIEAPP}, and AHIQ \cite{AHIQ}. As shown in Tab.\ref{FR_compare}, our proposed method performs in line with the current state-of-the-art model AHIQ, which indicates the effectiveness of our proposed unified architecture on IQA tasks. Moreover, predicted quality scores shown in Tab.\ref{pred_compare} further indicate our proposed YOTO performing well on both FR and NR tasks. The predicted IQA scores can correctly align with the MOS ranking based on HVS.

In addition to the above FR and NR benchmarks, we also trained YOTO on the PIPAL dataset. As shown in Tab.\ref{PIPAL_compare}, our proposed model with a Swin-Tiny backbone achieved on-par performance against the state-of-the-art MANIQA and AHIQ. Better scores were obtained when switching to a more powerful backbone, i.e. Swin-Small. Complexity-wise, the proposed YOTO has fewer parameters and MACs compared with the other state-of-the-art models, indicating the efficiency of our proposed method.

\textbf{Joint Training}
We additionally present the metrics attained through joint training of the model on both NR and FR datasets. Results are shown in \textcolor{red}{red} in Tab.\ref{NR_compare} and Tab.\ref{FR_compare}. \textbf{Our model achieves commensurate performance with the FR task trained alone, while exhibiting a notable advancement over the NR task trained alone.} This observation underscores a shared characteristic between FR and NR tasks, suggesting that our proposed unified framework adeptly captures relevant hidden patterns during the joint training, thereby further boosting the performance of NR task.

{\renewcommand{\arraystretch}{1.2}
\begin{table*}[t]
\centering
\setlength\tabcolsep{5pt}
\caption{Quantitaive comparison between our proposed NR-IQA network and other state-of-the-art NR-IQA methods. Postfix \textbf{-R} and \textbf{-S} stands for model trained on NR task only with ResNet50 and SWIN-Tiny Transformer encoder backbone, respectively. Model end with $\dag$ is jointly trained under FR \& NR tasks. Our jointly trained model achieves better performance on NR tasks.}
\begin{tabular}{c|cc|cc|cc|cc||cc|cc|cc} 
\bottomrule
\bottomrule
\multicolumn{1}{c|}{\multirow{2}{*}{\textbf{Methods}}}  & \multicolumn{2}{c|}{\textbf{LIVE}}         & \multicolumn{2}{c|}{\textbf{CSIQ}}          & \multicolumn{2}{c|}{\textbf{TID2013}}      & \multicolumn{2}{c||}{\textbf{KADID10K}}     & \multicolumn{2}{c|}{\textbf{LIVE-C}}      & \multicolumn{2}{c|}{\textbf{KonIQ10K}}      & \multicolumn{2}{c}{\textbf{LIVEFB}}\\
\multicolumn{1}{c|}{}                        & PLCC & \multicolumn{1}{c|}{SROCC} & PLCC & \multicolumn{1}{c|}{SROCC} & PLCC & \multicolumn{1}{c|}{SROCC} & PLCC &  \multicolumn{1}{c||}{SROCC} & PLCC & \multicolumn{1}{c|}{SROCC} & PLCC & \multicolumn{1}{c|}{SROCC} & PLCC & \multicolumn{1}{c}{SROCC}             \\ 
\hline
DIIVINE                  & 0.908 & 0.892 & 0.776 & 0.804 & 0.567 & 0.643 & 0.435 & 0.413 & 0.591 & 0.588 & 0.558 & 0.546 & 0.187 & 0.092 \\
BRISQUE                  & 0.944 & 0.929 & 0.748 & 0.812 & 0.571 & 0.626 & 0.567 & 0.528 & 0.629 & 0.629 & 0.685 & 0.681 & 0.341 & 0.303 \\
ILNIQE                   & 0.906 & 0.902 & 0.865 & 0.822 & 0.648 & 0.521 & 0.558 & 0.528 & 0.508 & 0.508 & 0.537 & 0.523 & 0.332 & 0.294 \\
BIECON                   & 0.961 & 0.958 & 0.823 & 0.815 & 0.762 & 0.717 & 0.648 & 0.623 & 0.613 & 0.613 & 0.654 & 0.651 & 0.428 & 0.407 \\
MEON                     & 0.955 & 0.951 & 0.864 & 0.852 & 0.824 & 0.808 & 0.691 & 0.604 & 0.710 & 0.697 & 0.628 & 0.611 & 0.394 & 0.365 \\
WaDIQaM                  & 0.955 & 0.960 & 0.844 & 0.852 & 0.855 & 0.835 & 0.752 & 0.739 & 0.671 & 0.682 & 0.807 & 0.804 & 0.467 & 0.455 \\
DBCNN                    & 0.971 & 0.968 & 0.959 & 0.946 & 0.865 & 0.816 & 0.856 & 0.851 & 0.869 & \textbf{0.869} & 0.884 & 0.875 & 0.551 & 0.545 \\
MetaIQA                  & 0.959 & 0.960 & 0.908 & 0.899 & 0.868 & 0.856 & 0.775 & 0.762 & 0.802 & 0.835 & 0.856 & 0.887 & 0.507 & 0.540 \\
P2P-BM                   & 0.958 & 0.959 & 0.902 & 0.899 & 0.856 & 0.862 & 0.849 & 0.840 & 0.842 & 0.844 & 0.885 & 0.872 & 0.598 & 0.526 \\
HyperIQA                 & 0.966 & 0.962 & 0.942 & 0.923 & 0.858 & 0.840 & 0.845 & 0.852 & 0.882 & 0.859 & 0.917 & 0.906 & 0.602 & 0.544 \\
TReS                     & 0.968 & 0.969 & 0.942 & 0.922 & 0.883 & 0.863 & 0.858 & 0.859 & 0.877 & 0.846 & \textbf{0.928} & 0.915 & 0.625 & 0.554 \\
\textbf{Ours-R}            & \textbf{0.976} & \textbf{0.974} & \textbf{0.962} & \textbf{0.952} & \textbf{0.908} & \textbf{0.902} & \textbf{0.884} & \textbf{0.885}  & \textbf{0.892}  & 0.841  & 0.925 & \textbf{0.917} & \textbf{0.637} & \textbf{0.563} \\
\bottomrule
\multicolumn{15}{c}{\textbf{Transformer-based}} \\
\bottomrule
TIQA                     & 0.965 & 0.949 & 0.838 & 0.825 & 0.858 & 0.846 & 0.855 & 0.850 & 0.861 & 0.845 & 0.903 & 0.892 & 0.581 & 0.541\\
MANIQA                   & 0.983 & 0.982 & 0.968 & 0.961 & 0.943 & 0.937 & 0.939 & 0.939 & - & - & - & - & - & - \\
\textbf{Ours-S}          & \textbf{0.986} & \textbf{0.986} & \textbf{0.970} & \textbf{0.964} & \textbf{0.956} & \textbf{0.953} & \textbf{0.942} & \textbf{0.941} & \textbf{0.903} & \textbf{0.869}  & \textbf{0.938} & \textbf{0.926} & \textbf{0.652} & \textbf{0.570} \\
\textbf{Ours-S\dag}          & \textcolor{red}{0.987} & \textcolor{red}{0.987} & \textcolor{red}{0.977} & \textcolor{red}{0.976} & \textcolor{red}{0.958} & \textcolor{red}{0.956} & \textcolor{red}{0.945} & \textcolor{red}{0.944} & - & - & - & - & - & - \\
\bottomrule
\bottomrule
\end{tabular}
\label{NR_compare}
\end{table*}
}

{\renewcommand{\arraystretch}{1.2}
\begin{table}[!t]
\centering
\setlength\tabcolsep{1.5pt}
\caption{Quantitaive comparison between YOTO and other FR methods in the field. $*$ stands for Transformer-based methods. Model end with $\dag$ is jointly trained under FR \& NR tasks while model end with $*$ is train on FR task only. Our jointly trained model achieved on-par performance of FR IQA.}
\begin{tabular}{c|cc|cc|cc|cc} 
\bottomrule
\bottomrule
\multicolumn{1}{c|}{\multirow{2}{*}{\textbf{Methods}}}  & \multicolumn{2}{c|}{\textbf{LIVE}}         & \multicolumn{2}{c|}{\textbf{CSIQ}}          & \multicolumn{2}{c|}{\textbf{TID2013}} & \multicolumn{2}{c}{\textbf{KADID10K}}\\
\multicolumn{1}{c|}{}  & PLCC & \multicolumn{1}{c|}{SROCC} & PLCC & \multicolumn{1}{c|}{SROCC} & PLCC & \multicolumn{1}{c|}{SROCC} & PLCC & \multicolumn{1}{c}{SROCC} \\ 
\hline
DOG                    & 0.966          & 0.963          & 0.943          & 0.954          & 0.934          & 0.926           & -      & - \\
DeepQA                 & 0.982          & 0.981          & 0.965          & 0.961          & 0.947          & 0.939           & 0.891  & 0.897 \\
DualCNN                & -              & -              & -              & -              & 0.924          & 0.926           & 0.949  & 0.941  \\
WaDIQaM                & 0.980          & 0.970          & -              & -              & 0.946          & 0.940           & 0.889  & 0.896 \\
PieAPP                 & 0.986          & 0.977          & 0.975          & 0.973          & 0.946          & 0.945           & -      & - \\
AHIQ*                  & \textbf{0.989} & 0.984          & 0.978          & 0.975          & \textbf{0.968} & 0.962           & -  & - \\
\textbf{Ours*}         & \textbf{0.989} & \textbf{0.988} &\textbf{0.979}  & \textbf{0.979} & 0.965          & \textbf{0.963}  & \textbf{0.950}  & \textbf{0.943}  \\
\textbf{Ours\dag}      & \textcolor{red}{0.988} & \textcolor{red}{0.988} &\textcolor{red}{0.978}  & \textcolor{red}{0.977} & \textcolor{red}{0.964} & \textcolor{red}{0.963}  & \textcolor{red}{0.950}  & \textcolor{red}{0.944}  \\
\bottomrule
\bottomrule
\end{tabular}
\label{FR_compare}
\end{table}
}

{\renewcommand{\arraystretch}{1.2}
\begin{table}[!t]
\centering
\setlength\tabcolsep{4pt}
\caption{Quantitaive comparison between YOTO and SOTA networks on the PIPAL test dataset, with FLOPs and parameters indicated. The suffix -T and -S represents Swin tiny and small backbones. Our YOTO achieved great performance with much fewer parameters and MACs, indicating the effectiveness of the proposed architecture.}
\begin{tabular}{c|c|c||cc|cc} 
\bottomrule
\bottomrule
\multicolumn{1}{c|}{\multirow{2}{*}{\textbf{Methods}}}  & \multicolumn{1}{c|}{\textbf{Params.}}  & \multicolumn{1}{c||}{\textbf{MAC}} & \multicolumn{2}{c|}{\textbf{PIPAL (NR)}}         & \multicolumn{2}{c}{\textbf{PIPAL (FR)}} \\
\multicolumn{1}{c|}{}  &(M) &(G) & PLCC & \multicolumn{1}{c|}{SROCC} & PLCC & \multicolumn{1}{c}{SROCC} \\ 
\hline
MANIQA                 & 127.73          & 108.62           & 0.667          & 0.702          & -              & -              \\
AHIQ                   & 132.62           & 168.98          & -              & -              & 0.823          & 0.813          \\
Ours-T        & 76.64            & 80.56           & 0.662          & 0.696          & 0.818          & 0.807
      \\
Ours-S        & \textbf{105.36}  & \textbf{98.62}  & \textbf{0.669}&\textbf{0.703} &\textbf{0.826} &\textbf{0.815} \\
\bottomrule
\bottomrule
\end{tabular}
\label{PIPAL_compare}
\end{table}
}
{\renewcommand{\arraystretch}{1.2}
\begin{table*}[!t]
\centering
\setlength\tabcolsep{1pt}
\caption{Quality score comparison for YOTO in FR and NR mode against other state-of-the-art models AHIQ (FR) and MANIQA (NR). Images are chosen from the TID2013 test set with Block Occlusion distortion type. From left to right: the pristine image and 5 distorted images with descending distortion levels. Different color codes are applied for better visualization of the \textcolor{red}{1st}, \textcolor{orange}{2nd}, \textcolor{green}{3rd}, \textcolor{cyan}{4th}, and \textcolor{blue}{5th} ranking of prediction scores.}

\begin{tabular}{c|c|c|c|c|c}
\bottomrule
\bottomrule
 \begin{minipage}{.16\linewidth}
    \includegraphics[width=\linewidth]{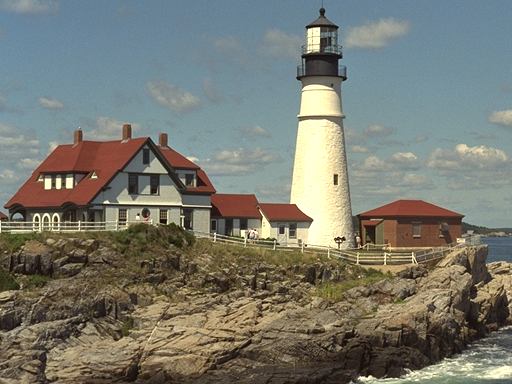}
 \end{minipage} &
  \begin{minipage}{.16\linewidth}
    \includegraphics[width=\linewidth]{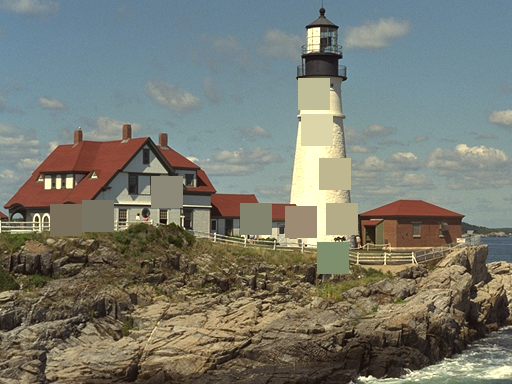}
 \end{minipage} &
  \begin{minipage}{.16\linewidth}
    \includegraphics[width=\linewidth]{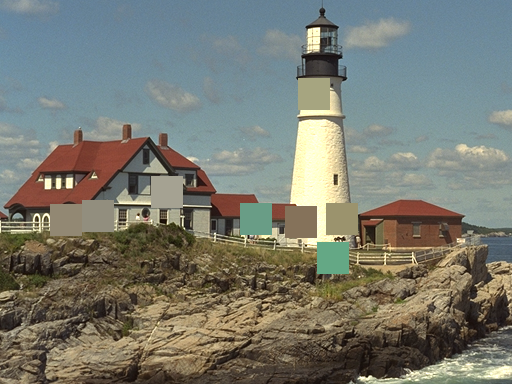}
 \end{minipage} &
  \begin{minipage}{.16\linewidth}
    \includegraphics[width=\linewidth]{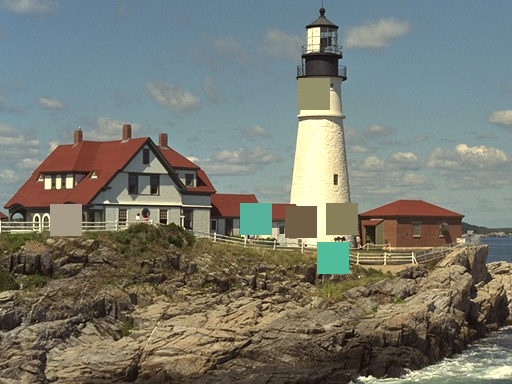}
 \end{minipage} &
  \begin{minipage}{.16\linewidth}
    \includegraphics[width=\linewidth]{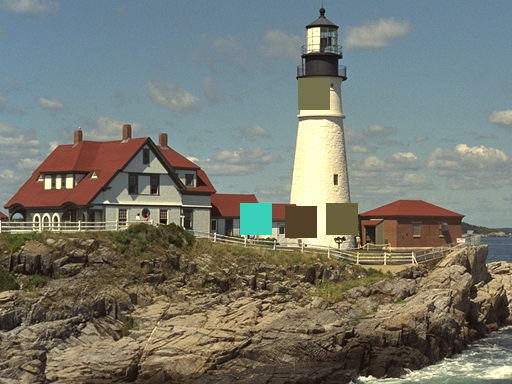}
 \end{minipage} &
  \begin{minipage}{.16\linewidth}
    \includegraphics[width=\linewidth]{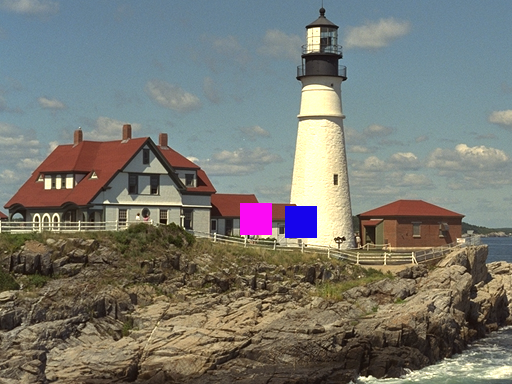}
 \end{minipage} \\
 Reference Image & Distortion 1 & Distortion 2 & Distortion 3 & Distortion 4 & Distortion 5 \\
 \hline
 MOS (normed)    &\textcolor{blue}{0.3959 (5th)} & \textcolor{cyan}{0.4181 (4th)} & \textcolor{green}{0.4382 (3rd)} & \textcolor{orange}{0.4742 (2nd)} & \textcolor{red}{0.5566 (1st)} \\
 YOTO (FR)      &\textcolor{blue}{0.4890 (5th)} & \textcolor{cyan}{0.4912 (4th)} & \textcolor{green}{0.4939 (3rd)} & \textcolor{orange}{0.5031 (2nd)} & \textcolor{red}{0.5771 (1st)} \\
 YOTO (NR)      &\textcolor{blue}{0.4823 (5th)} & \textcolor{cyan}{0.4937 (4th)} & \textcolor{green}{0.5000 (3rd)} & \textcolor{orange}{0.5308 (2nd)} & \textcolor{red}{0.5636 (1st)} \\
 AHIQ (FR)       &\textcolor{blue}{0.4901 (5th)} & \textcolor{cyan}{0.4945 (4th)} & \textcolor{green}{0.5104 (3rd)} & \textcolor{red}{0.5713 (1st)}    & \textcolor{orange}{0.5628 (2nd)} \\
 MANIQA (NR)      &\textcolor{cyan}{0.5112 (4th)} & \textcolor{blue}{0.4996 (5th)} & \textcolor{orange}{0.5644 (2nd)}& \textcolor{green}{0.5364 (3rd)}  & \textcolor{red}{0.5709 (1st)} \\
\bottomrule
\bottomrule
 \begin{minipage}{.16\linewidth}
    \includegraphics[width=\linewidth]{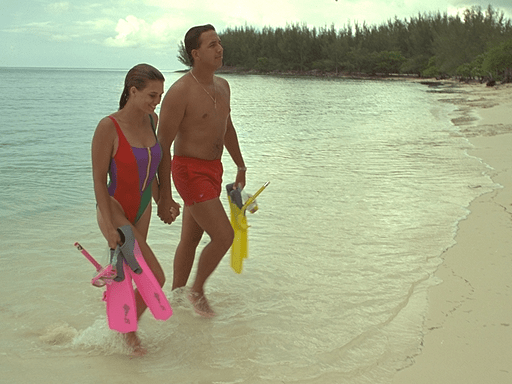}
 \end{minipage} &
  \begin{minipage}{.16\linewidth}
    \includegraphics[width=\linewidth]{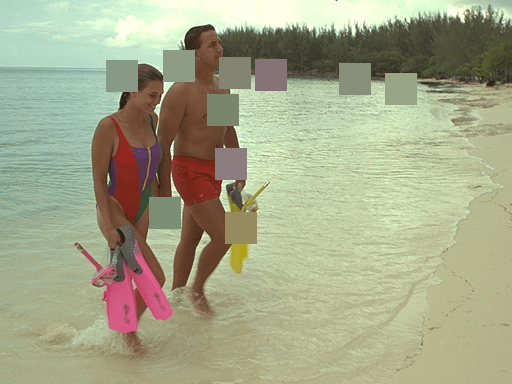}
 \end{minipage} &
  \begin{minipage}{.16\linewidth}
    \includegraphics[width=\linewidth]{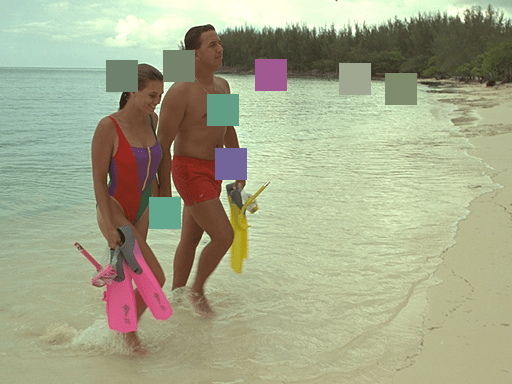}
 \end{minipage} &
  \begin{minipage}{.16\linewidth}
    \includegraphics[width=\linewidth]{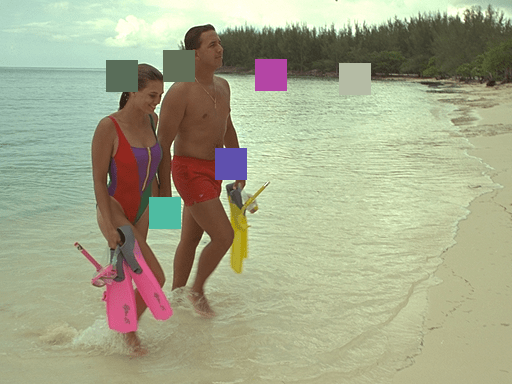}
 \end{minipage} &
  \begin{minipage}{.16\linewidth}
    \includegraphics[width=\linewidth]{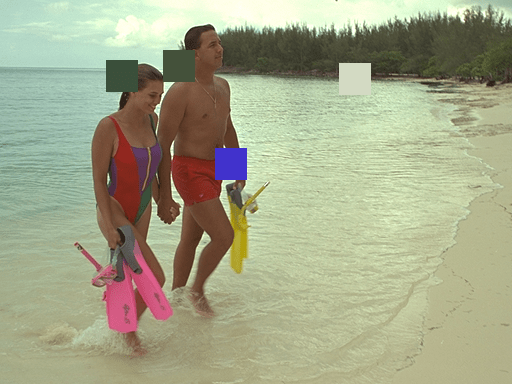}
 \end{minipage} &
  \begin{minipage}{.16\linewidth}
    \includegraphics[width=\linewidth]{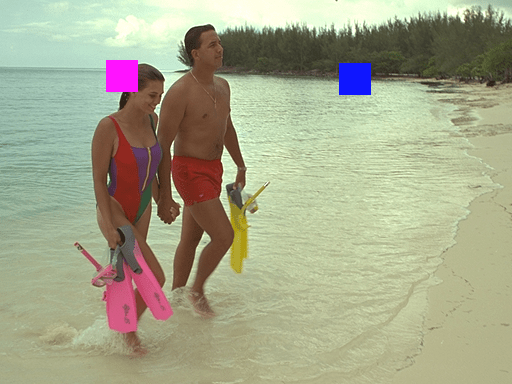}
 \end{minipage} \\
 Reference Image & Distortion 1 & Distortion 2 & Distortion 3 & Distortion 4 & Distortion 5 \\
\hline
 MOS (normed)    &\textcolor{blue}{0.4400 (5th)} &\textcolor{cyan}{0.4435 (4th)} &\textcolor{green}{0.4574 (3rd)} &\textcolor{orange}{0.4948 (2nd)} &\textcolor{red}{0.5379 (1st)} \\
 YOTO (FR)      &\textcolor{blue}{0.4242 (5th)} &\textcolor{cyan}{0.4930 (4th)} &\textcolor{green}{0.5056 (3rd)} &\textcolor{orange}{0.5280 (2nd)} &\textcolor{red}{0.6165 (1st)} \\
 YOTO (NR)      &\textcolor{blue}{0.4111 (5th)} &\textcolor{cyan}{0.4532 (4th)} &\textcolor{green}{0.4757 (3rd)} &\textcolor{orange}{0.5616 (2nd)} &\textcolor{red}{0.6124 (1st)} \\
 AHIQ (FR)       &\textcolor{blue}{0.4694 (5th)} &\textcolor{green}{0.5102 (3rd)} &\textcolor{cyan}{0.4811 (4th)} &\textcolor{red}{0.5709 (1st)} &\textcolor{orange}{0.5598 (2nd)} \\
 MANIQA (NR)      &\textcolor{cyan}{0.4776 (4th)} &\textcolor{blue}{0.4380 (5th)} &\textcolor{green}{0.5083 (3rd)} &\textcolor{red}{0.5702 (1st)} &\textcolor{orange}{0.5349 (2nd)} \\
 
\bottomrule
\bottomrule
\end{tabular}

\label{pred_compare}
\end{table*}
}

\section{Ablation Studies}
\textbf{Effectiveness of the Proposed Modules}
We reconfigured our network by switching on/off HA and SDA respectively and trained the network under TID2013 and CSIQ for both FR and NR IQA tasks. Evaluation results are listed in Tab.\ref{hsa_csca}. Compared with HA, the SDA module is more effective on IQA tasks. This is probably due to its effectiveness in modeling semantic damage across different stages, further indicating semantic damage is an important factor of IQA. When equipped with both models, the network yields the best performance. More detailed illustrations of feature maps from HA and SDA will be discussed in \textit{Sec. Visualization of HA and SDA Module} of ablation studies.

\textbf{Effectiveness of Hierarchical Attention}
In order to compare the effectiveness of the HA module, we fixed the scale factor $r$ to 1 and stacked the attention layer 2,4, and 6 times. Experiment results are listed in Tab.\ref{HA_ablation}. In comparison to 6 layers of plain attention, two layers of HA with scale factors 1 and 2 produce competitive results. The HA module is efficient in computation and effective in performance. Besides, as the scale factor increases, the improvement of HA becomes marginal. This is probably because a large scale factor $r$ will generate small attention blocks, which result in similar scores in each block and thus hinder the performance.

\begin{figure*}[!ht]
\centering
\footnotesize
\stackunder[3pt]{\includegraphics[width=.11\linewidth]{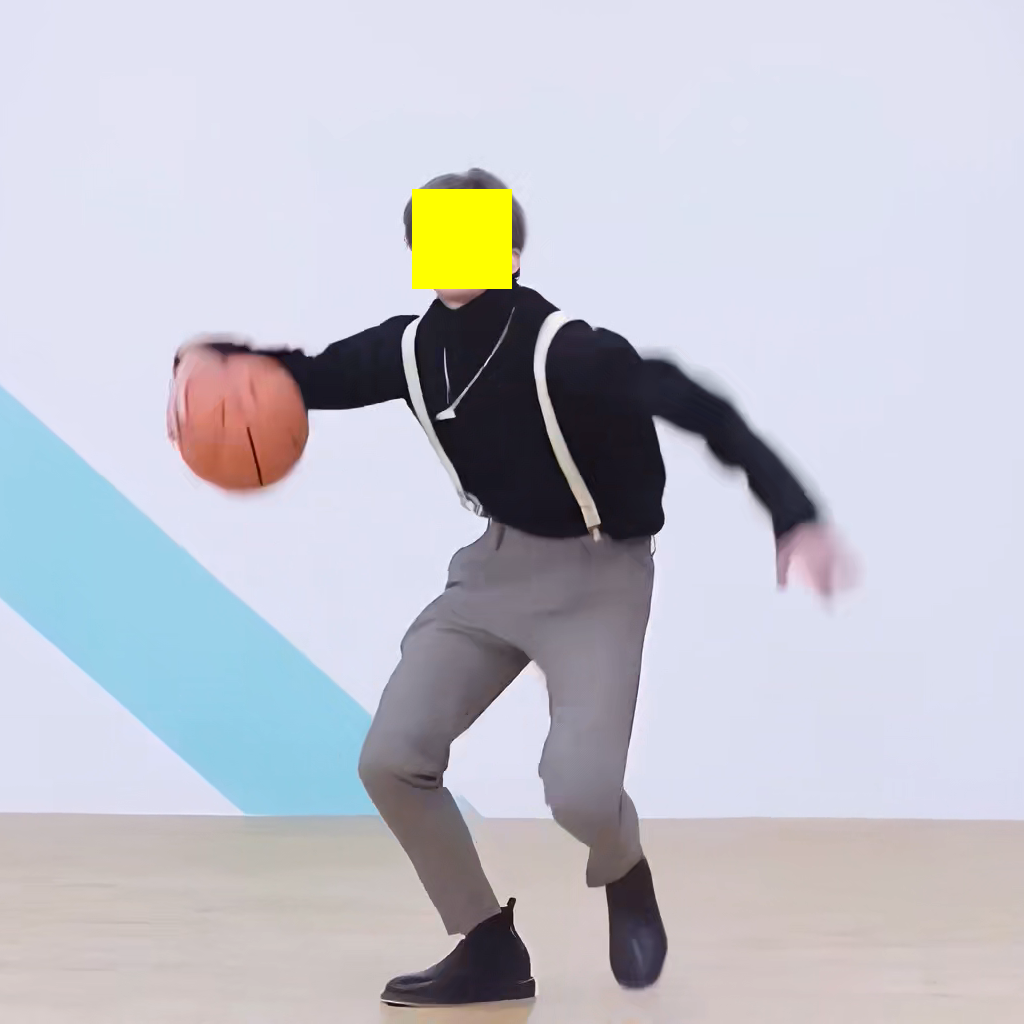}}{Distorted Image}%
\stackunder[3pt]{\includegraphics[width=.125\linewidth]{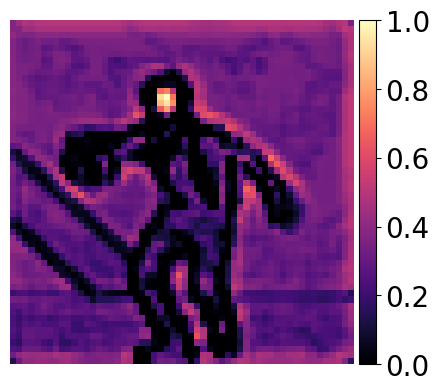}}{Encoder Stg.1}%
\stackunder[3pt]{\includegraphics[width=.125\linewidth]{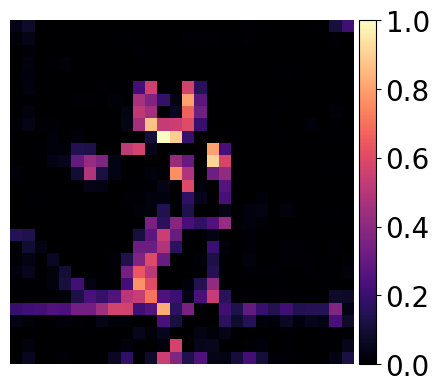}}{Encoder Stg.2}%
\stackunder[3pt]{\includegraphics[width=.125\linewidth]{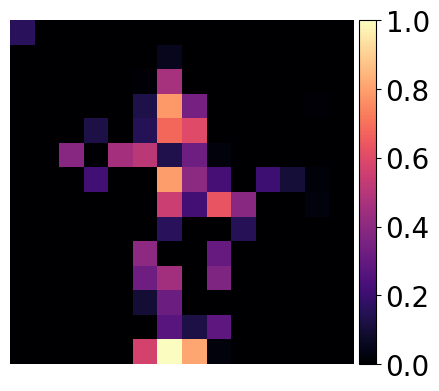}}{Encoder Stg.3}%
\stackunder[3pt]{\includegraphics[width=.125\linewidth]{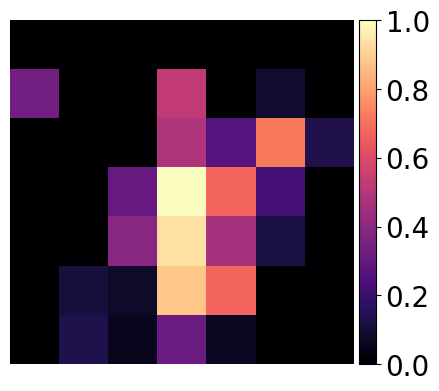}}{Encoder Stg.4}%
\stackunder[3pt]{\includegraphics[width=.125\linewidth]{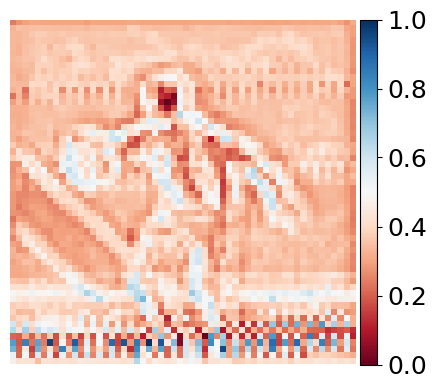}}{SDA(1,2)}%
\stackunder[3pt]{\includegraphics[width=.125\linewidth]{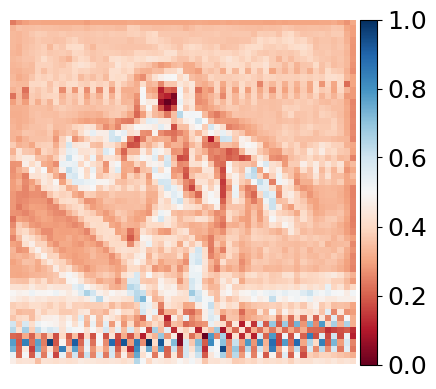}}{SDA(1,3)}%
\stackunder[3pt]{\includegraphics[width=.125\linewidth]{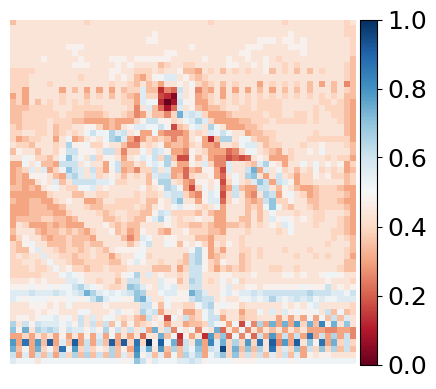}}{SDA(1,4)}%

\stackunder[3pt]{\includegraphics[width=.11\linewidth]{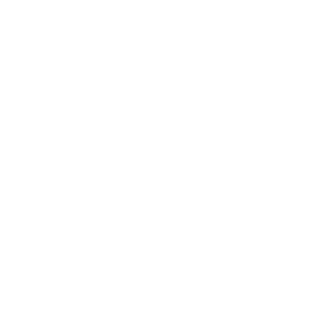}}{}%
\stackunder[3pt]{\includegraphics[width=.125\linewidth]{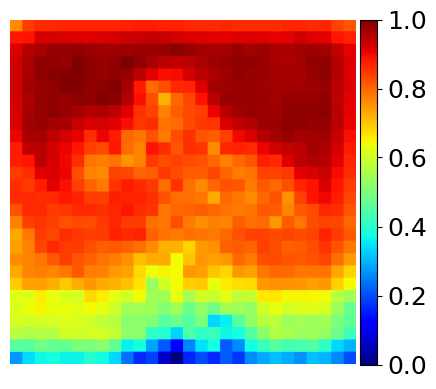}}{HA(1)}%
\stackunder[3pt]{\includegraphics[width=.125\linewidth]{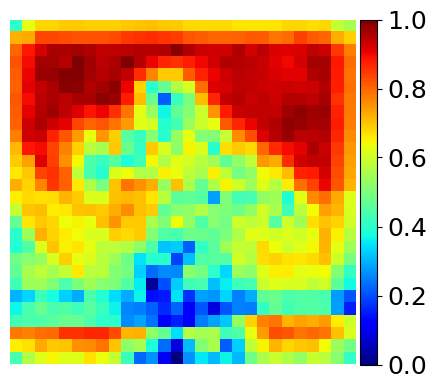}}{HA(2)}%
\stackunder[3pt]{\includegraphics[width=.125\linewidth]{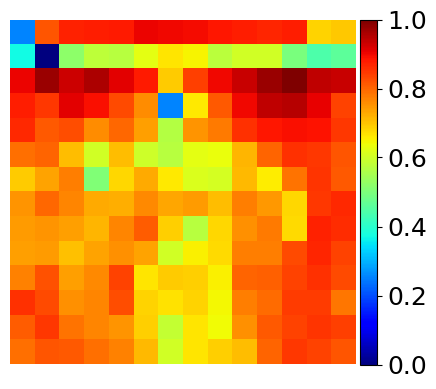}}{HA(3)}%
\stackunder[3pt]{\includegraphics[width=.125\linewidth]{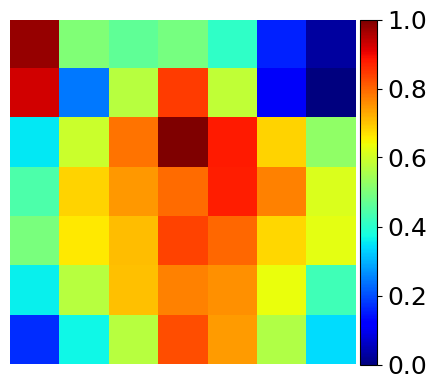}}{HA(4)}%
\stackunder[3pt]{\includegraphics[width=.125\linewidth]{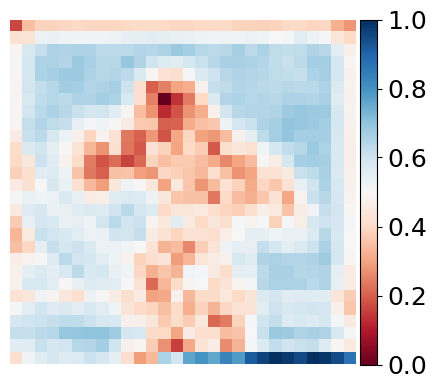}}{SDA(2,3)}%
\stackunder[3pt]{\includegraphics[width=.125\linewidth]{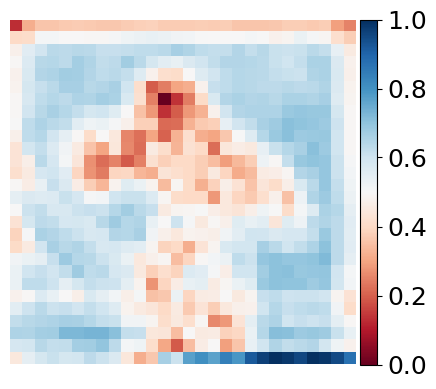}}{SDA(2,4)}%
\stackunder[3pt]{\includegraphics[width=.125\linewidth]{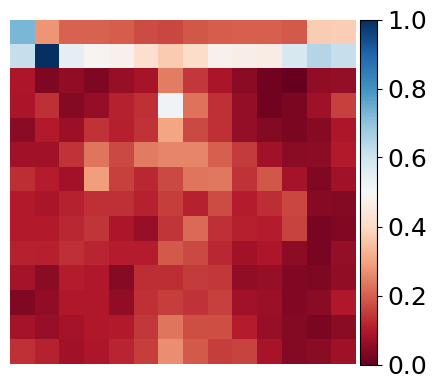}}{SDA(3,4)}%

\stackunder[3pt]{\includegraphics[width=.11\linewidth]{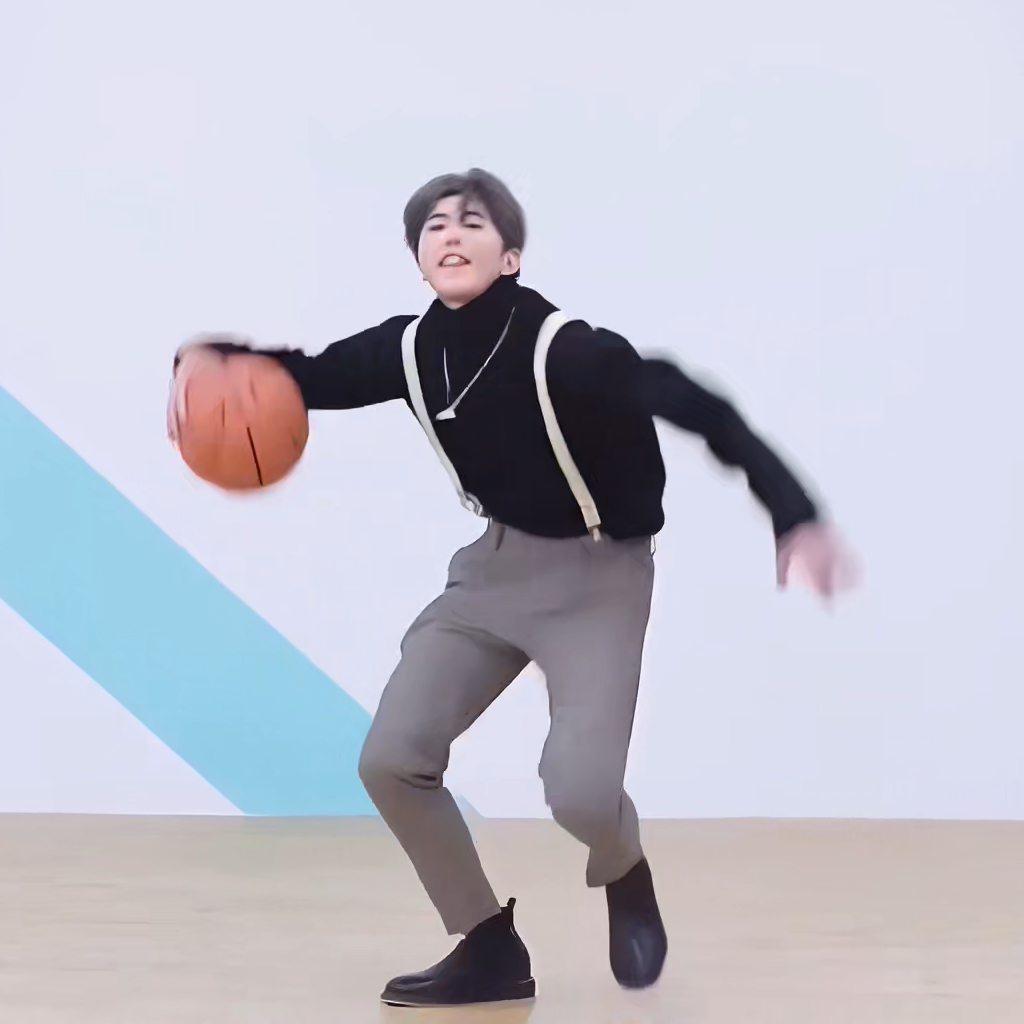}}{Reference Image}%
\stackunder[3pt]{\includegraphics[width=.125\linewidth]{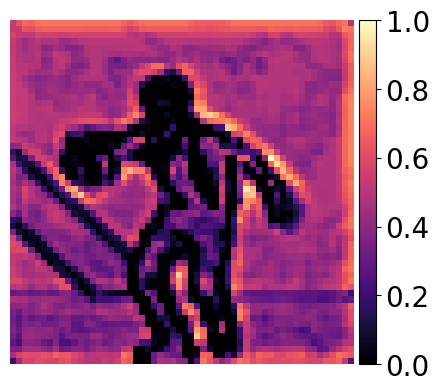}}{Encoder Stg.1}%
\stackunder[3pt]{\includegraphics[width=.125\linewidth]{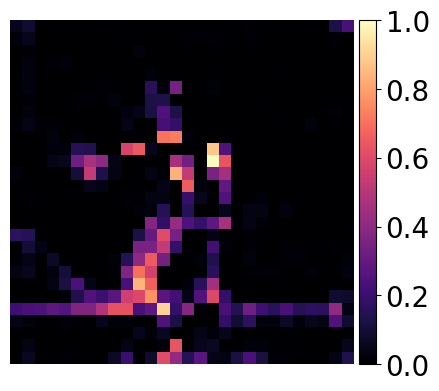}}{Encoder Stg.2}%
\stackunder[3pt]{\includegraphics[width=.125\linewidth]{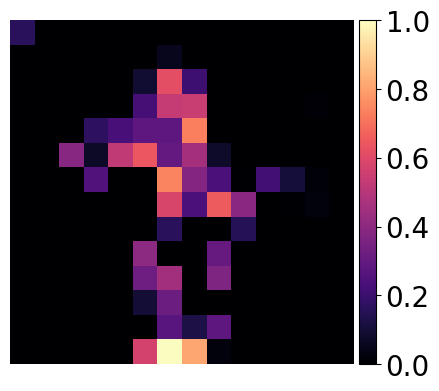}}{Encoder Stg.3}%
\stackunder[3pt]{\includegraphics[width=.125\linewidth]{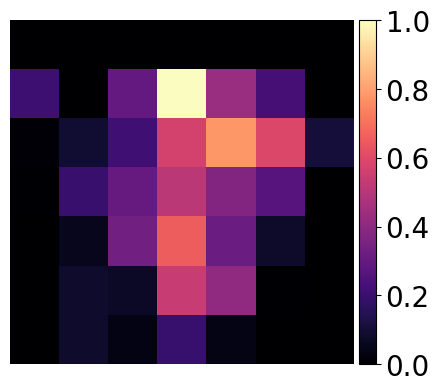}}{Encoder Stg.4}%
\stackunder[3pt]{\includegraphics[width=.125\linewidth]{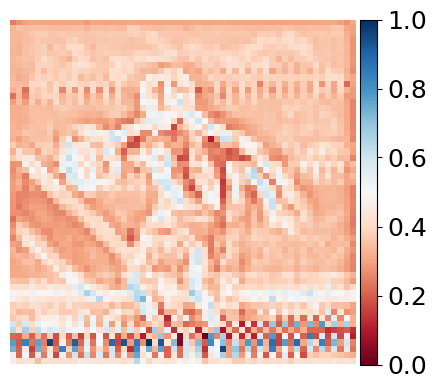}}{SDA(1,2)}%
\stackunder[3pt]{\includegraphics[width=.125\linewidth]{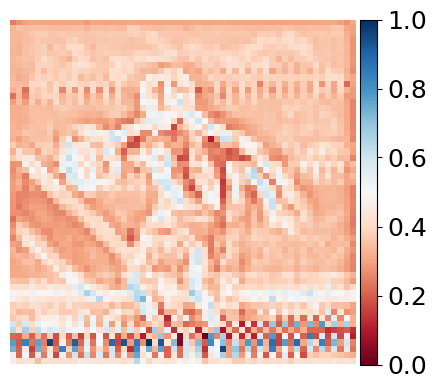}}{SDA(1,3)}%
\stackunder[3pt]{\includegraphics[width=.125\linewidth]{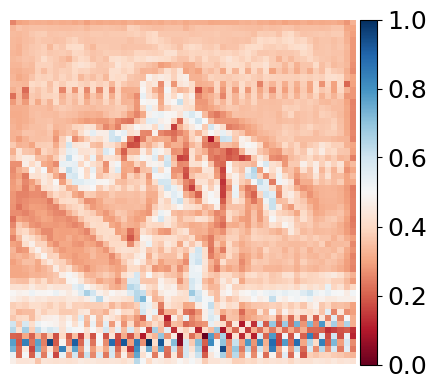}}{SDA(1,4)}%

\stackunder[3pt]{\includegraphics[width=.11\linewidth]{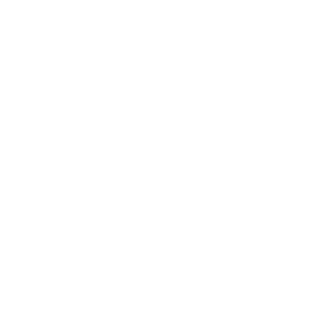}}{}%
\stackunder[3pt]{\includegraphics[width=.125\linewidth]{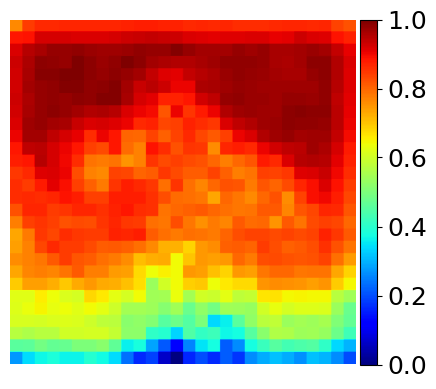}}{HA(1)}%
\stackunder[3pt]{\includegraphics[width=.125\linewidth]{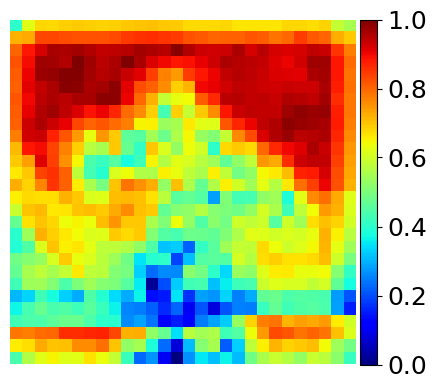}}{HA(2)}%
\stackunder[3pt]{\includegraphics[width=.125\linewidth]{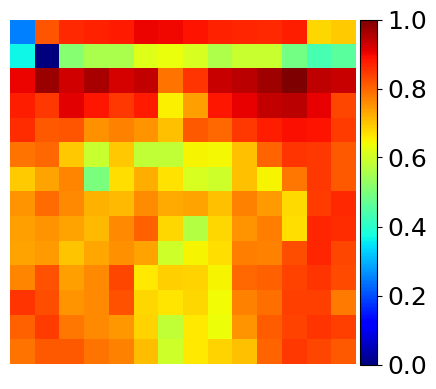}}{HA(3)}%
\stackunder[3pt]{\includegraphics[width=.125\linewidth]{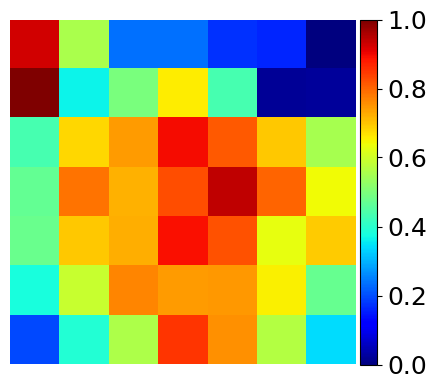}}{HA(4)}%
\stackunder[3pt]{\includegraphics[width=.125\linewidth]{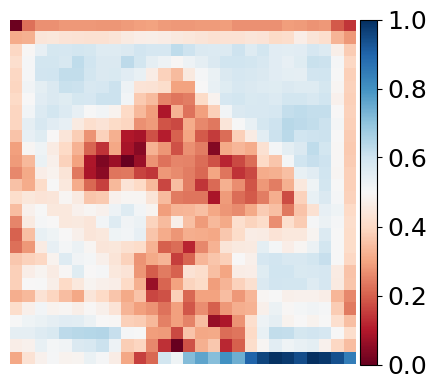}}{SDA(2,3)}%
\stackunder[3pt]{\includegraphics[width=.125\linewidth]{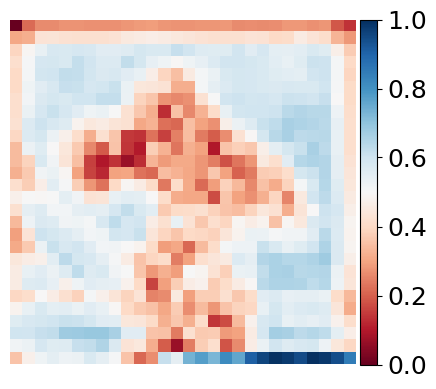}}{SDA(2,4)}%
\stackunder[3pt]{\includegraphics[width=.125\linewidth]{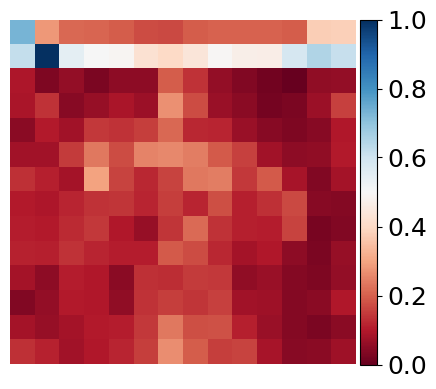}}{SDA(3,4)}%

\caption{Illustration of how YOTO detects semantic impact when \textbf{Block Occlusion} distortion is presented in a test image in the wild. Features for both distorted and reference images are listed for ease of comparison and a better understanding of each proposed module. ResNet50 is adopted as the encoder backbone. Top half: distorted image and features from encoder stages 1 to 4, HA module, and SDA module are listed respectively. SDA(1,2) denotes cross-attentioned features between encoder stages 1 and 2. Bottom half: feature maps retrieved at the same position in encoder stages for reference images. It is clear that self-attention is effective in identifying spatial important regions, but lacks consistency across different encoder stages. On the other hand, cross-attention can effectively and consistently reflect similarities between features from different encoder stages, which is beneficial in indicating the residual effect of distortion.}
\label{kunkun_vis}
\end{figure*}

\begin{figure*}[!h]
\centering
\footnotesize
\stackunder[3pt]{\includegraphics[width=.11\linewidth]{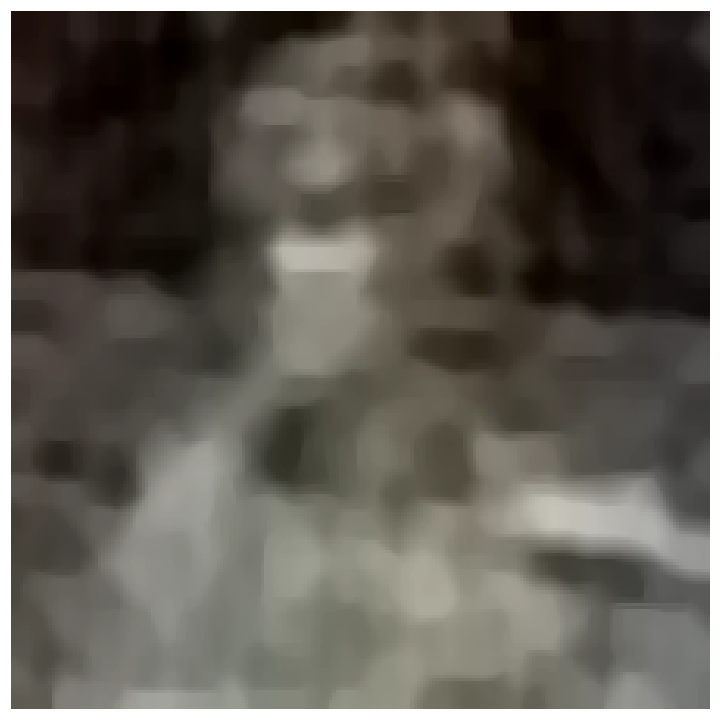}}{Most Distorted}%
\stackunder[3pt]{\includegraphics[width=.125\linewidth]{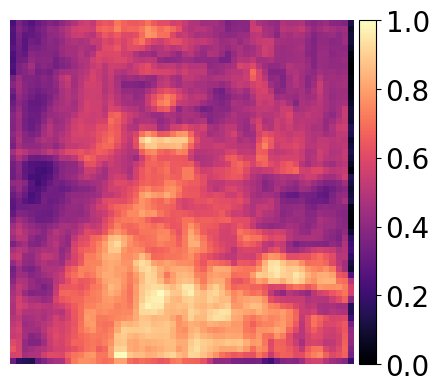}}{Encoder Stg.1}%
\stackunder[3pt]{\includegraphics[width=.125\linewidth]{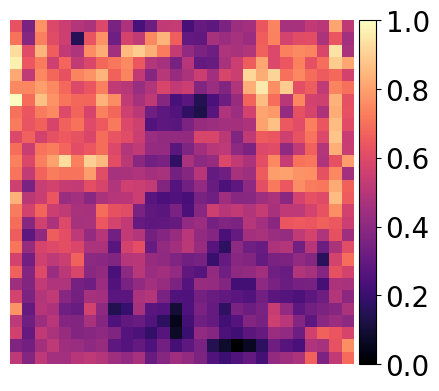}}{Encoder Stg.2}%
\stackunder[3pt]{\includegraphics[width=.125\linewidth]{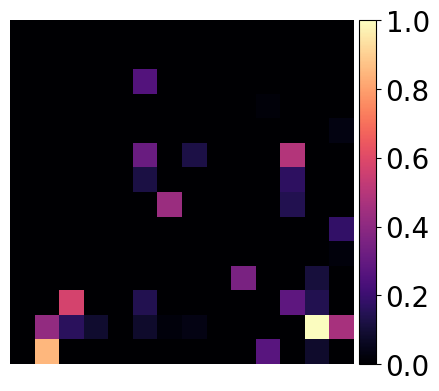}}{Encoder Stg.3}%
\stackunder[3pt]{\includegraphics[width=.125\linewidth]{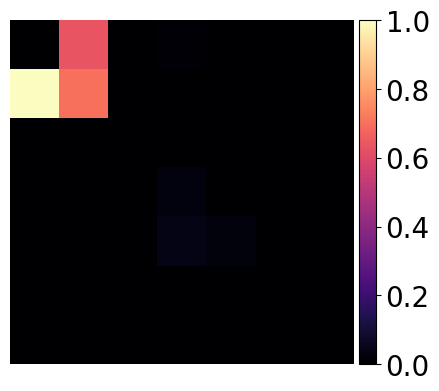}}{Encoder Stg.4}%
\stackunder[3pt]{\includegraphics[width=.125\linewidth]{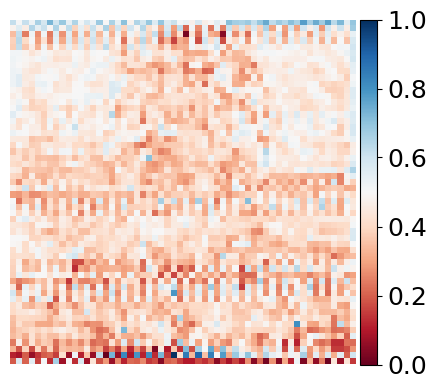}}{SDA(1,2)}%
\stackunder[3pt]{\includegraphics[width=.125\linewidth]{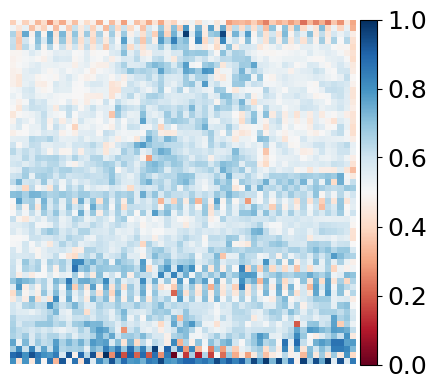}}{SDA(1,3)}%
\stackunder[3pt]{\includegraphics[width=.125\linewidth]{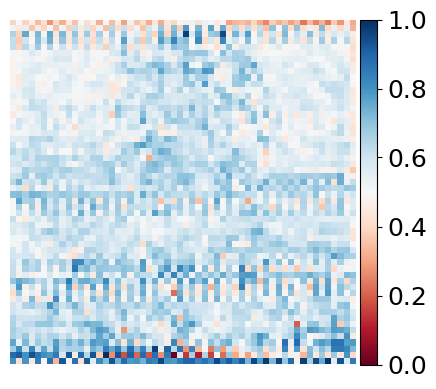}}{SDA(1,4)}%

\stackunder[3pt]{\includegraphics[width=.11\linewidth]{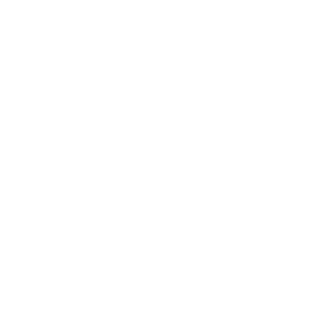}}{}%
\stackunder[3pt]{\includegraphics[width=.125\linewidth]{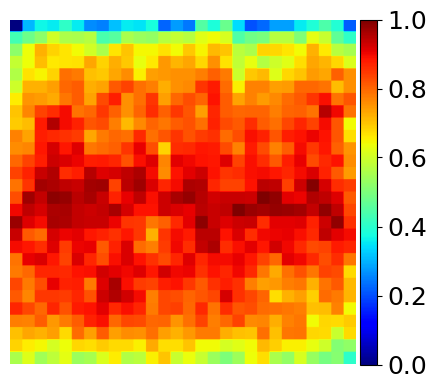}}{HA(1)}%
\stackunder[3pt]{\includegraphics[width=.125\linewidth]{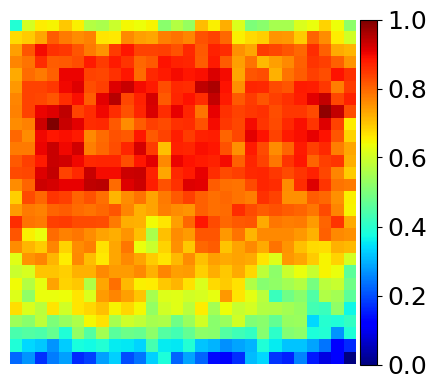}}{HA(2)}%
\stackunder[3pt]{\includegraphics[width=.125\linewidth]{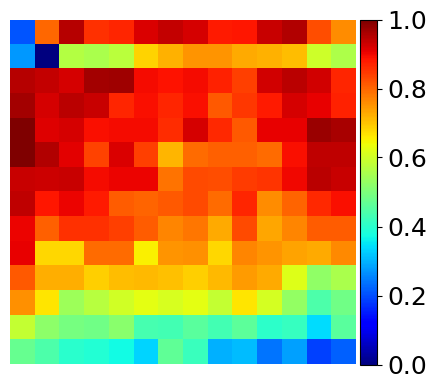}}{HA(3)}%
\stackunder[3pt]{\includegraphics[width=.125\linewidth]{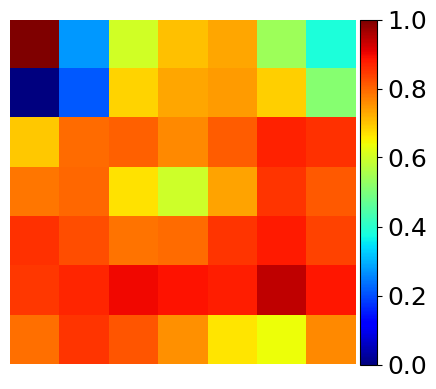}}{HA(4)}%
\stackunder[3pt]{\includegraphics[width=.125\linewidth]{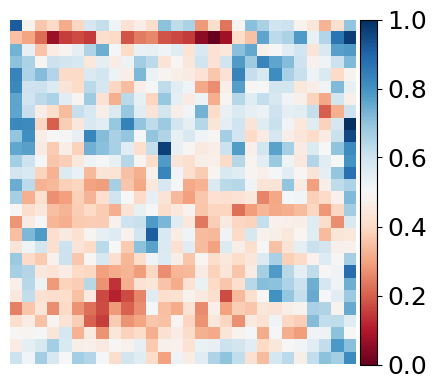}}{SDA(2,3)}%
\stackunder[3pt]{\includegraphics[width=.125\linewidth]{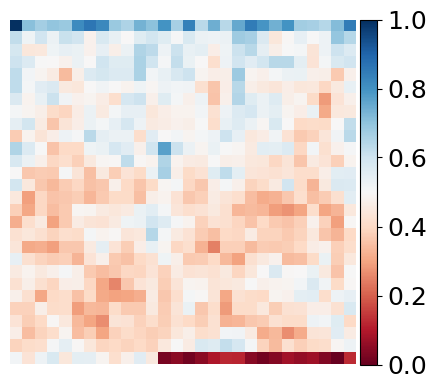}}{SDA(2,4)}%
\stackunder[3pt]{\includegraphics[width=.125\linewidth]{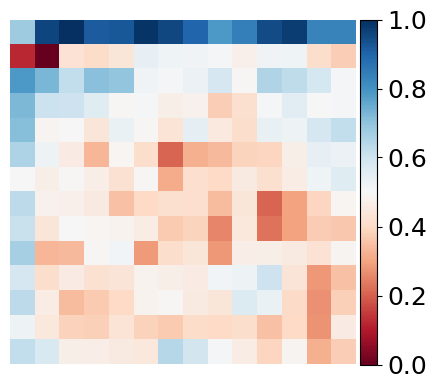}}{SDA(3,4)}%

\stackunder[3pt]{\includegraphics[width=.11\linewidth]{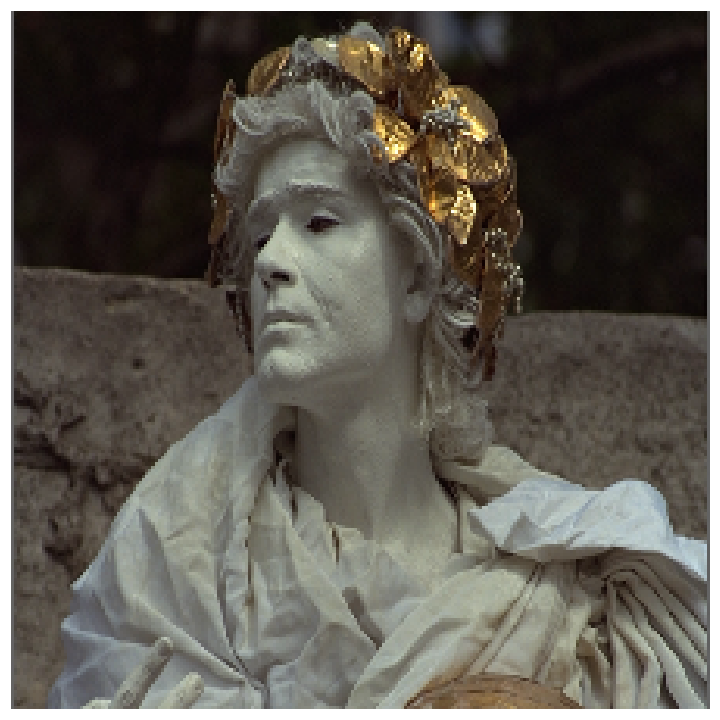}}{Least Distorted}%
\stackunder[3pt]{\includegraphics[width=.125\linewidth]{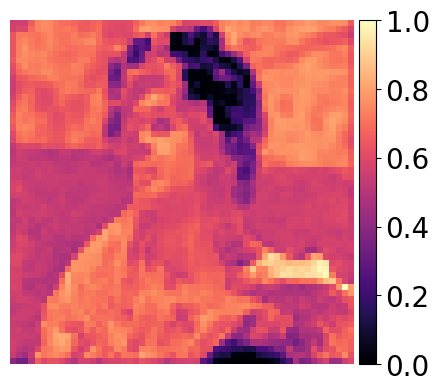}}{Encoder Stg.1}%
\stackunder[3pt]{\includegraphics[width=.125\linewidth]{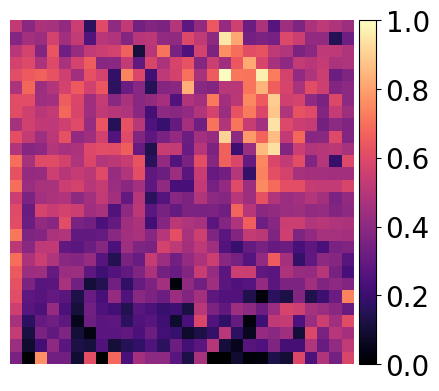}}{Encoder Stg.2}%
\stackunder[3pt]{\includegraphics[width=.125\linewidth]{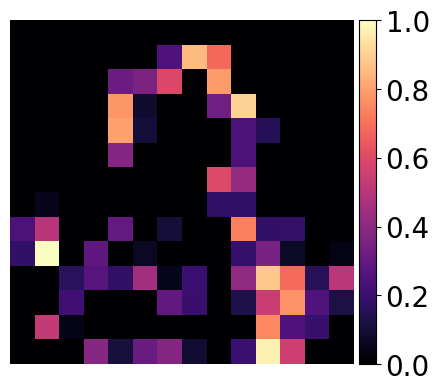}}{Encoder Stg.3}%
\stackunder[3pt]{\includegraphics[width=.125\linewidth]{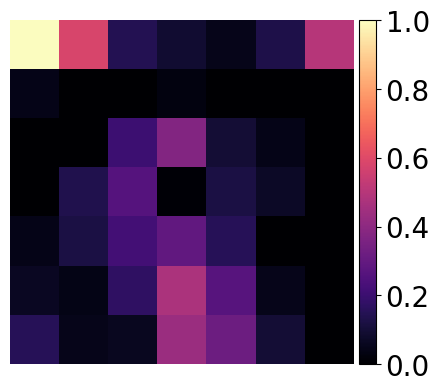}}{Encoder Stg.4}%
\stackunder[3pt]{\includegraphics[width=.125\linewidth]{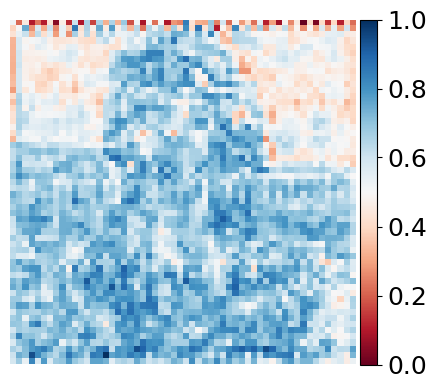}}{SDA(1,2)}%
\stackunder[3pt]{\includegraphics[width=.125\linewidth]{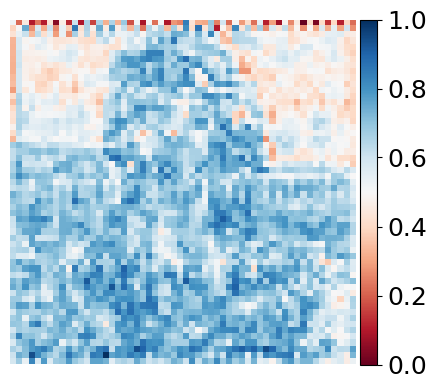}}{SDA(1,3)}%
\stackunder[3pt]{\includegraphics[width=.125\linewidth]{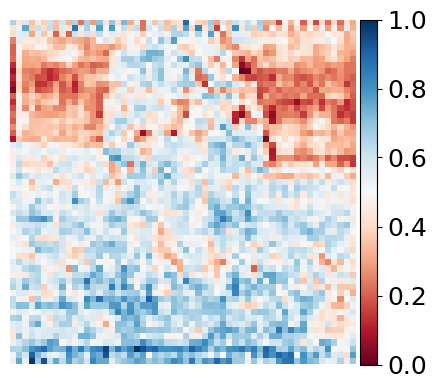}}{SDA(1,4)}%

\stackunder[3pt]{\includegraphics[width=.11\linewidth]{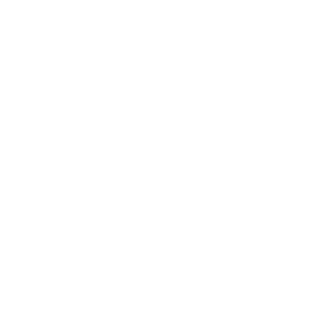}}{}%
\stackunder[3pt]{\includegraphics[width=.125\linewidth]{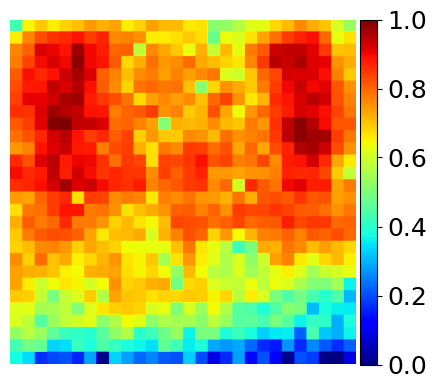}}{HA(1)}%
\stackunder[3pt]{\includegraphics[width=.125\linewidth]{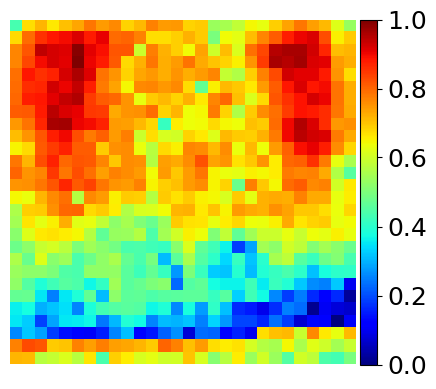}}{HA(2)}%
\stackunder[3pt]{\includegraphics[width=.125\linewidth]{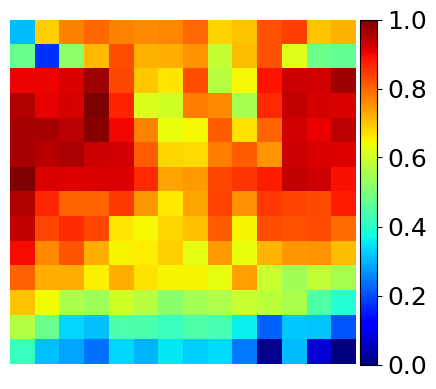}}{HA(3)}%
\stackunder[3pt]{\includegraphics[width=.125\linewidth]{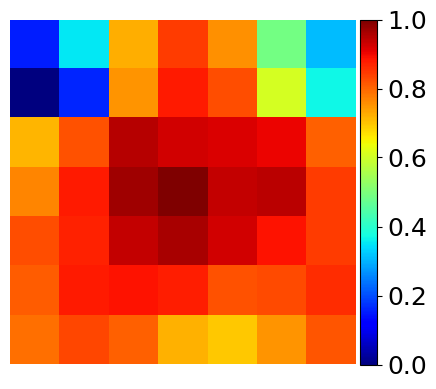}}{HA(4)}%
\stackunder[3pt]{\includegraphics[width=.125\linewidth]{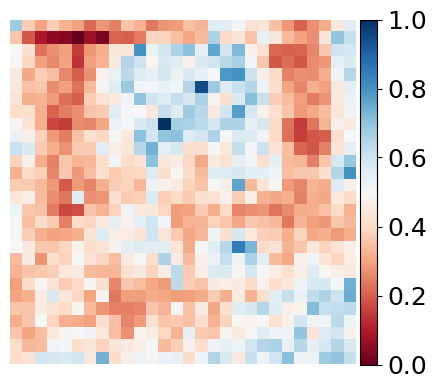}}{SDA(2,3)}%
\stackunder[3pt]{\includegraphics[width=.125\linewidth]{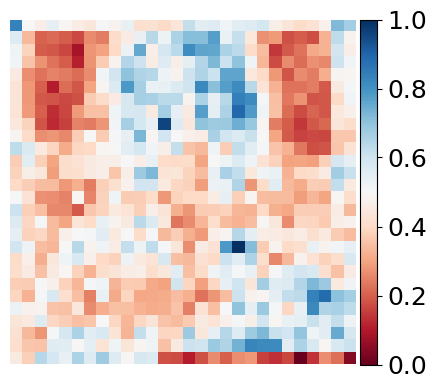}}{SDA(2,4)}%
\stackunder[3pt]{\includegraphics[width=.125\linewidth]{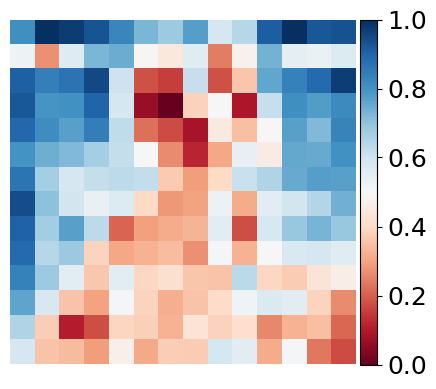}}{SDA(3,4)}%

\caption{Illustration of how YOTO detects semantic impact when different levels of \textbf{Sparse Sampling} distortion is presented in a test image from TID2013. Similar to Fig. \ref{kunkun_vis}, features for the most distorted (level 5) and the least distorted (level 1) images are listed for ease of comparison and a better understanding of each proposed module. ResNet50 is adopted as the encoder backbone. Top half: distortion level 5 and features from encoder stages 1 to 4, HA module, and SDA module are listed respectively. Bottom half: feature maps retrieved at the same position in encoder stages for the least distorted image. As shown above, cross-attention is effective in determining correlations between encoder stages and estimating the semantic impact caused by distortion.}
\label{dis5_1_comp}
\end{figure*}

\textbf{Visualization of HA and SDA Module}
\label{vis}
We investigate the effectiveness of our proposed HA and SDA module by visualizing encoder features and the attention matrix learned. ResNet50 version of YOTO is used, as shown in Fig.\ref{kunkun_vis} and Fig.\ref{dis5_1_comp}. We choose two typical distortion patterns in this analysis, specifically, Block Occlusion Distortion and Sparse Sampling. Feature maps are sampled at the same channel position of each encoder layer for both reference images and distorted images. Visualizations reflect the fact that mild distortion will be filtered out by the shallow encoder layers while severe distortion will persist and ultimately affect the deepest layer of features. The proposed HA and SDA modules can effectively reflect the amount of spatial distortion and the correlation between shallow and deep features, which can help the network estimate quality scores.

\textbf{Cross-dataset Evaluation}
Following prior research \cite{maniqa, AHIQ}, we also perform cross-dataset evaluation by training our model on KADID-10K and LIVE and testing on CSIQ and TID2013 respectively for both FR and NR tasks. As shown in Tab.\ref{cross_dataset}, our proposed network achieves acceptable generalization ability on both FR and NR tasks.

\textbf{Mix-dataset Joint Training}
A more comprehensive evaluation is conducted by jointly training our model on different FR and NR datasets. We combined one authentic NR dataset and two synthetic FR datasets to train our model and evaluate it under NR IQA. The network will receive FR image pairs $(distorted, reference)$ or NR image pairs $(distorted, distorted)$ depending on the sampling. For NR models like MANIQA and TReS, only NR image pairs are loaded. The train/test split ratio is 0.8 for each dataset and we train the network for 200 epochs with a learning rate of $1e^{-4}$. The experiment results shown in Tab.\ref{mix_dataset} demonstrate that our model is still able to achieve fairly good performance and outperform other single-task models with \textbf{up to 10\% performance increment} using this joint training strategy with a mixed dataset.

{\renewcommand{\arraystretch}{1.2}
\begin{table*}[!t]
\centering
\caption{Ablation study on the effectiveness of HA and SDA module on FR and NR tasks.}
\begin{tabular}{ccc|cc|cc||cc|cc} 
\bottomrule
\bottomrule
\multicolumn{3}{c|}{\textbf{Configuration}} & \multicolumn{2}{c|}{\textbf{TID2013 (NR)}} & \multicolumn{2}{c||}{\textbf{CSIQ(NR)}} & \multicolumn{2}{c|}{\textbf{TID2013 (FR)}} & \multicolumn{2}{c}{\textbf{CSIQ(FR)}}  \\
HA & SDA (no cone) & SDA (cone)                   & PLCC & SRCC                       & PLCC & SRCC                   & PLCC & SRCC                       & PLCC & SRCC                   \\ 
\hline
\CheckmarkBold   &                  & & 0.946 & 0.943 & 0.961 & 0.952 & 0.953 & 0.949 & 0.966 & 0.968                   \\ 
\cline{4-11}
                 &  \CheckmarkBold  & & 0.944 & 0.944 & 0.960 & 0.950 & 0.950 & 0.948 & 0.964 & 0.965                   \\ 
\cline{4-11}
                 &  & \CheckmarkBold  & 0.949 & 0.947 & 0.966 & 0.958 & 0.958 & 0.955 & 0.971 & 0.972                   \\ 
\cline{4-11}
\CheckmarkBold   &  & \CheckmarkBold  & \textbf{0.956} & \textbf{0.953} & \textbf{0.970} & \textbf{0.964} & \textbf{0.965} & \textbf{0.963} &\textbf{0.979} & \textbf{0.979}                  \\ 
\bottomrule
\bottomrule
\end{tabular}
\label{hsa_csca}
\end{table*}
}
{\renewcommand{\arraystretch}{1.2}
\begin{table*}[!t]
\centering
\setlength\tabcolsep{1.5pt}
\caption{Effectiveness of the HA module. Left half: by fixing the scale factor $r=1$, HA becomes plain attention. Multiple HA layers are stacked and their results are reported. Right half: HA module with different scale factor settings. As shown below, HA(1,2) performs better than 6 layers of plain attention.}
\begin{tabular}{c|cc|cc||c|cc|cc} 
\bottomrule
\bottomrule
\multirow{2}{*}{\begin{tabular}[c]{@{}c@{}}\textbf{HA(r=1)}\\No. Layers\end{tabular}} & \multicolumn{2}{c|}{\textbf{TID2013}} & \multicolumn{2}{c||}{\textbf{CSIQ}} & \multirow{2}{*}{\textbf{HA(r)}} & \multicolumn{2}{c|}{\textbf{TID2013}} & \multicolumn{2}{c}{\textbf{CSIQ}}  \\
        & PLCC & SROCC   & PLCC & SROCC    &               & PLCC & SROCC    & PLCC & SROCC  \\ 
\hline
2       & 0.948 &0.941    & 0.953 &0.948   & HA(1,2)      & 0.956 & 0.953   & 0.970 & 0.964 \\
4       & 0.951 &0.946    & 0.958 &0.951   & HA(1,2,4)    & 0.953 & 0.948   & 0.962 & 0.955 \\
6       & 0.953 &0.949    & 0.966 &0.975   & HA(1,2,4,8)  & 0.950 & 0.946   & 0.957 & 0.949 \\
\bottomrule
\bottomrule
\end{tabular}
\label{HA_ablation}
\end{table*}
}
{\renewcommand{\arraystretch}{1.2}
\begin{table*}[!ht]
\centering
\setlength\tabcolsep{7pt}
\caption{Mix-dataset evaluation. The model was trained on the mixed training dataset of \{LIVE, CSIQ, KONIQ-10K\} and \{TID2013, KADID-10K, LIVE-C\}, then evaluated on corresponding testing datasets under \textbf{NR IQA mode}. \textcolor{blue}{\textbf{Blue}} and \textcolor{green}{\textbf{Green}} denote \textcolor{blue}{\textbf{NR}} and \textcolor{green}{\textbf{FR}} datasets. Our proposed YOTO achieves outstanding performance on both FR and NR joint-training tasks.}
\begin{tabular}{c|cc|cc|cc||cc|cc|cc} 
\bottomrule
\bottomrule
\textbf{Train}                 & \multicolumn{6}{c||}{\textbf{\textcolor{blue}{KONIQ} + \textcolor{green}{LIVE} + \textcolor{green}{CSIQ}}}  & \multicolumn{6}{c}{\textbf{\textcolor{blue}{LIVEC} + \textcolor{green}{TID2013} + \textcolor{green}{KADID}}}    \\
\hline
\multirow{2}{*}{\textbf{Test}} & \multicolumn{2}{c|}{\textbf{KONIQ}}  & \multicolumn{2}{c|}{\textbf{LIVE~}} & \multicolumn{2}{c||}{\textbf{CSIQ}} & \multicolumn{2}{c|}{\textbf{LIVEC}}  & \multicolumn{2}{c|}{\textbf{TID2013}} & \multicolumn{2}{c}{\textbf{KADID}}\\
                               & PLCC & SROCC   & PLCC & SROCC      & PLCC & SROCC  & PLCC & SROCC   & PLCC & SROCC      & PLCC & SROCC \\ 
\hline
Tres                          & 0.764 & 0.773 & 0.842 & 0.839   & 0.816 & 0.810      & 0.702 & 0.709 & 0.748 & 0.731   & 0.734 & 0.719\\
\hline
MANIQA                        & 0.828 & 0.817  & 0.903 & 0.898  & 0.885 & 0.856      & 0.871 & 0.822  & 0.889 & 0.882  & 0.914 & 0.915\\
\hline
Ours-T                        
& \textbf{0.866} 
& \textbf{0.853} 
& \textbf{0.971} 
& \textbf{0.972}   
& \textbf{0.956} 
& \textbf{0.944}

& \textbf{0.912} 
& \textbf{0.871} 
& \textbf{0.924} 
& \textbf{0.923}   
& \textbf{0.929} 
& \textbf{0.933}\\
\bottomrule
\bottomrule
\end{tabular}
\label{mix_dataset}
\end{table*}
}

{\renewcommand{\arraystretch}{1.2}
\begin{table*}[!ht]
\centering
\setlength\tabcolsep{10pt}
\caption{Cross-dataset evaluation. Part of the results are borrowed from \cite{AHIQ}. $*$ stands for Transformer-based methods. The proposed network demonstrates good generalization ability.}
\begin{tabular}{c|cc|cc||c|cc|cc} 
\bottomrule
\bottomrule
\textbf{Mode}               & \multicolumn{4}{c||}{\textbf{FR}}                                     &\textbf{Mode}               & \multicolumn{4}{c}{\textbf{\textbf{NR}}}  \\ 
\hline
\multirow{2}{*}{\textbf{Train/Test}} & \multicolumn{2}{c|}{\textbf{LIVE\ /\ CSIQ}} & \multicolumn{2}{c||}{\textbf{KADID\ /\ TID2013}} & \multirow{2}{*}{\textbf{Train/Test}} & \multicolumn{2}{c|}{\textbf{LIVE\ /\ CSIQ}} & \multicolumn{2}{c}{\textbf{KADID\ /\ TID2013}} \\
& SROCC & PLCC                   & SROCC & PLCC                       &                            & SROCC & PLCC                   & SROCC & PLCC\\  

\hline
WaDIQaM            & 0.909   & 0.895           & 0.834 & 0.831                  &TReS             & 0.652 & 0.680             & 0.664 & 0.639                  \\
\hline
PieAPP             & 0.882   & 0.876           & 0.859 & 0.876                  &MANIQA*          & \textbf{0.788} & 0.850             & 0.753 & \textbf{0.763}                  \\ 
\hline

\textbf{Ours-FR*}  & \textbf{0.917}   & \textbf{0.908}           & \textbf{0.885} & \textbf{0.889}                  &\textbf{Ours-NR*}& 0.782 & \textbf{0.851}             & \textbf{0.755} & 0.762                  \\ 
\bottomrule
\bottomrule
\end{tabular}
\label{cross_dataset}
\end{table*}
}

\section{Discussion}
\textbf{Necessity of a Unified Model} Human perception of image quality relies significantly on the interaction of various modalities of information, such as visual, textual, and auditory cues. Recent research endeavors have begun integrating additional modalities, such as audio \cite{audio_visual,mmsod}, into this domain. Hence, in the long term, IQA models are envisioned to integrate diverse modalities within a unified architecture. Our work represents an initial step in this direction by unifying NR and FR IQA. This integration not only reduces the current disparities between FR and NR IQA models but also lays the groundwork for a scalable backbone for future multimodal endeavors.

\textbf{Extension to More Modalities} The inclusion of multiple modalities not only enriches the model with additional information but also brings it closer to approximating authentic human experiences in real-life situations. Existing works predominantly engage in the extraction of features from diverse modalities, followed by feature fusion through methods such as hand-crafted rules, pooling, or neural networks to estimate quality scores, as mentioned in \cite{av_survey}.

\begin{figure}[!t]
\centering
\includegraphics[width=\linewidth]{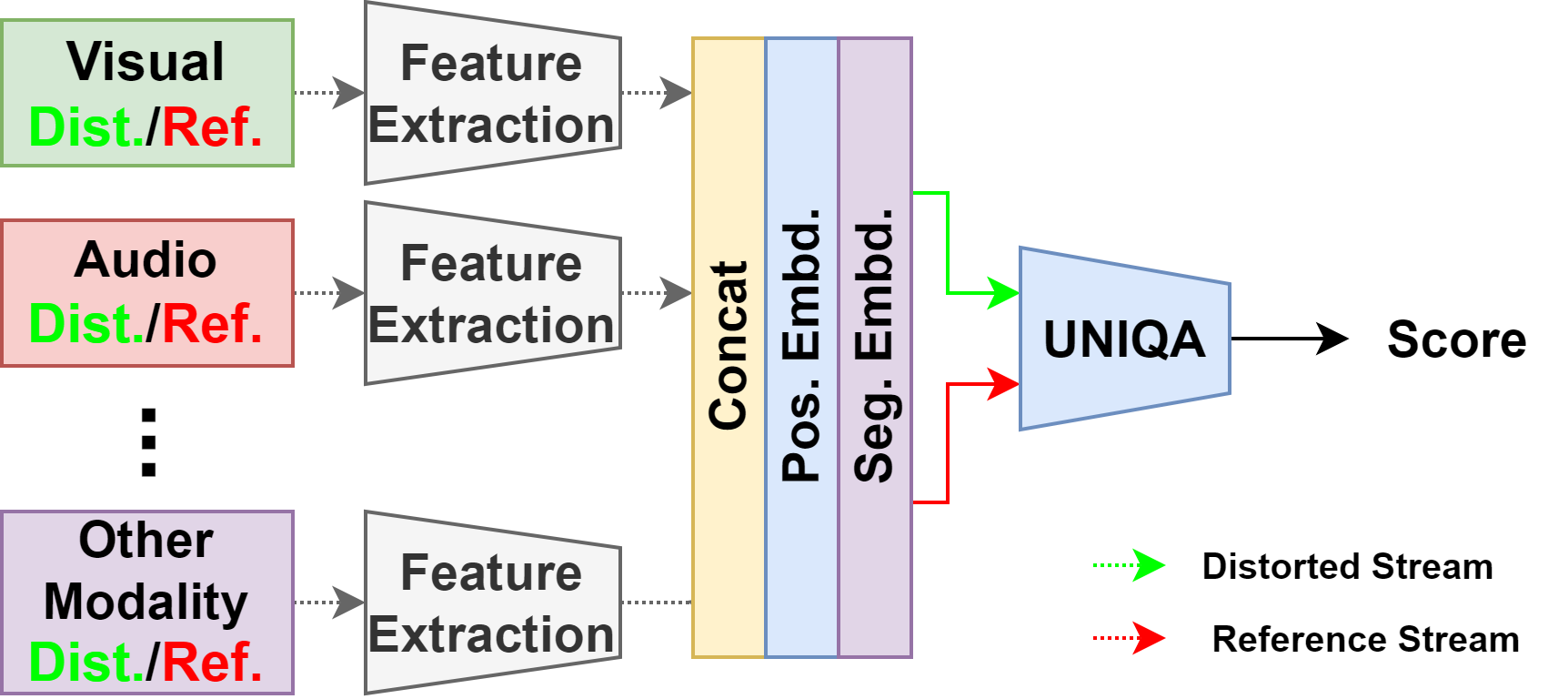}
\caption{An illustration of a possible architecture for YOTO when multiple modalities are present for image quality assessment. Feature extraction networks need to be implemented for modality-specific feature extraction. Features from each modality should be treated like RGB images for ease of channel-wise concatenation. After applying position embedding and segment-wise embedding, the proposed YOTO will then perform multi-scale self/cross-attention for quality score prediction.}
\label{mmYOTO}
\end{figure}

The architecture proposed in this work has the potential to handle multi-modal scenarios. Specifically, for audio-visual IQA, a feasible approach involves feature extraction for each modality for both reference and distorted data, as shown in Fig.\ref{mmYOTO}. Features are concatenated channel-wise and then are fed into YOTO. By utilizing self-attention, the importance of intra-modality features is computed, while cross-attention is employed to ascertain the alignment of inter-modality features. The final step involves leveraging a multi-layer perceptron (MLP) network to estimate quality scores. The actual performance and ablations of the proposed multi-modal architecture await investigation.

\textbf{Extension to More Applications} In recent years, the application of IQA has expanded into various domains, including but not limited to Virtual Reality IQA, Light Field IQA, and Screen Content IQA, as mentioned in surveys \cite{survey1, survey2}. Despite the diverse nature of these application scenarios, the fundamental approach to addressing IQA challenges remains remarkably consistent. Broadly speaking, in FR tasks, the core strategy involves a meticulous examination of the disparities between a reference image and its distorted counterpart \cite{fr_struc, fr_gauss, fr_edge}. In NR scenarios, the emphasis shifts towards a nuanced understanding of the inherent data distribution \cite{nr_stat, nr_spa}, entailing the modeling of data consistency and other pertinent features \cite{nr_feat_re, region_feat} to holistically evaluate image quality.

The proposed YOTO appears to be promising in terms of its potential for various IQA domains. In the context of traditional IQA and Screen Content IQA, distorted and reference images typically manifest as individual pairs. The proposed YOTO is capable of seamlessly extending to both FR and NR scenarios in such tasks. However, YOTO might not be suitable for Light Field IQA since the prevalent format often involves a collection of images, such as Sub-Aperture Image (SAI), Epipolar Plane Image (EPI), or Micro-lens Array Coded Photograph (MACPI).

\section{Conclusion}
In this work, we have proposed a novel architecture based on the transformer and attention mechanism to integrate the FR and NR IQA tasks. To address the input discrepancy between the FR and NR tasks, we introduce a Hierarchical Attention (HA) module that dynamically switches between attention types based on the presence of reference images. Additionally, we partition the attention matrix into blocks at different layers of the HA module to further explore local attention and achieve faster convergence. Considering different types of distortion exhibit varying levels of residual across encoder layers, we devise a Semantic Distortion Aware (SDA) module to capture the similarity between features of different encoder stages. Feature pyramid is split into cones and cross-attention is applied across different layers of each cone. The experimental results demonstrate that our unified framework is capable of handling both FR and NR tasks and achieves overall state-of-the-art results on their respective benchmarks by only being trained once, indicating the effectiveness of the proposed architecture.

\bibliographystyle{ieee_fullname}
\bibliography{main}

\end{document}